\documentclass{article}
\usepackage{natbib}
\usepackage[dblblindworkshop, final]{neurips_2026}
\workshoptitle{Interpretability as a Science}

\usepackage{graphicx}
\usepackage{amsmath}
\usepackage{placeins}
\usepackage{float}
\usepackage{flafter}
\usepackage[utf8]{inputenc}
\usepackage[T1]{fontenc}
\usepackage{hyperref}
\usepackage{url}
\usepackage{booktabs}
\usepackage{amsfonts}
\usepackage{nicefrac}
\usepackage{microtype}
\usepackage{xcolor}
\usepackage{multirow}
\usepackage[capitalize,noabbrev]{cleveref}
\usepackage{xspace}

\hypersetup{
  pdftitle={Detection Is Not Early Warning: Evaluating Token-Level Alarms on Language Model Failures},
  pdfauthor={Marco De Negri, Luca Sartori, Lorenzo Baggi, Moritz Reihs, Eduard Frank Durech, Mennatallah El-Assady, Valentina Pyatkin, Anna Hedstr\"{o}m},
}

\providecommand{\sd}[1]{{\scriptstyle\pm#1}}
\newcommand{\method}{\textsc{Prewarn}\xspace}

\title{Detection Is Not Early Warning: Evaluating Token-Level Alarms on Language Model Failures}

\author{%
  Marco De Negri$^{1}$ \enspace
  Luca Sartori$^{1}$ \enspace
  Lorenzo Baggi$^{1}$ \enspace
  Moritz Reihs$^{1}$ \\
  \bfseries Eduard Frank \v{D}urech$^{2}$ \enspace
  Mennatallah El-Assady$^{1}$ \enspace
  Valentina Pyatkin$^{1, 2}$ \enspace
  Anna Hedstr\"{o}m$^{2,3}$ \\[0.6em]
  {\normalfont
  $^{1}$ETH Zurich \quad
  $^{2}$ETH AI Center \quad
  $^{3}$Solaris Research}
}

\begin{document}

\maketitle

\begin{abstract}
Many failures of language models begin at a specific token, the onset. A monitor that reads the model during generation can classify the finished response, detect the failure once it is under way, or warn before it begins, and only a warning leaves time to stop or steer the generation. Monitors are usually evaluated with response-level metrics such as AUROC, which ignore when the monitor fires. We introduce \method{}, an evaluation protocol for any failure with an annotated onset. It calibrates each monitor's threshold on validation data to an error rate chosen for the application, runs the monitor, and scores its alarms relative to the onset on all three tasks. As a case study we apply it to repetitive degeneration in three open language models, with early warning as the main task, using linear probes on activations and six untrained baselines. Response-level metrics saturate on this failure, and a rule that reads only the response length scores almost perfectly, while probes that AUROC ranks as equal warn very differently under \method{}. Probes warn before the onset on most degenerate responses of two of the three models, low-rank adapters below the probe more than double the share of tokens just before the onset that it flags, largely by flagging degenerate responses long before the onset, and no untrained baseline warns much more often than random alarms. We release the probes with calibrated thresholds at \href{https://huggingface.co/collections/marcodene/prewarn-degeneration-probes}{\nolinkurl{huggingface.co/collections/marcodene/prewarn-degeneration-probes}}.
\end{abstract}

\section{Introduction}
\label{sec:introduction}

Many failures of language models begin at a specific token, the \emph{onset}. From that token on, a response may, for example, turn harmful \citep{li2025fineharm}, make unsupported claims \citep{niu2024ragtruth}, or fall into a repetition loop that runs until the token budget is exhausted \citep{holtzman2020curious,welleck2020neural,pipis2025wait}. A monitor that reads the model during generation can serve three tasks: \emph{response classification} gives one verdict on the finished response, \emph{failure detection} raises an alarm once the failure is under way, and \emph{early warning} raises one before it begins. Only an early warning leaves time to stop, steer or resample the generation before any of the failure is produced \citep{li2025fineharm,hedstrom2025mera,zhang2025backtracking}. We ask how such monitors should be evaluated, focusing on early warning, with repetitive degeneration as the case study.

\begin{figure}[h]
    \centering
    \includegraphics[width=1\linewidth]{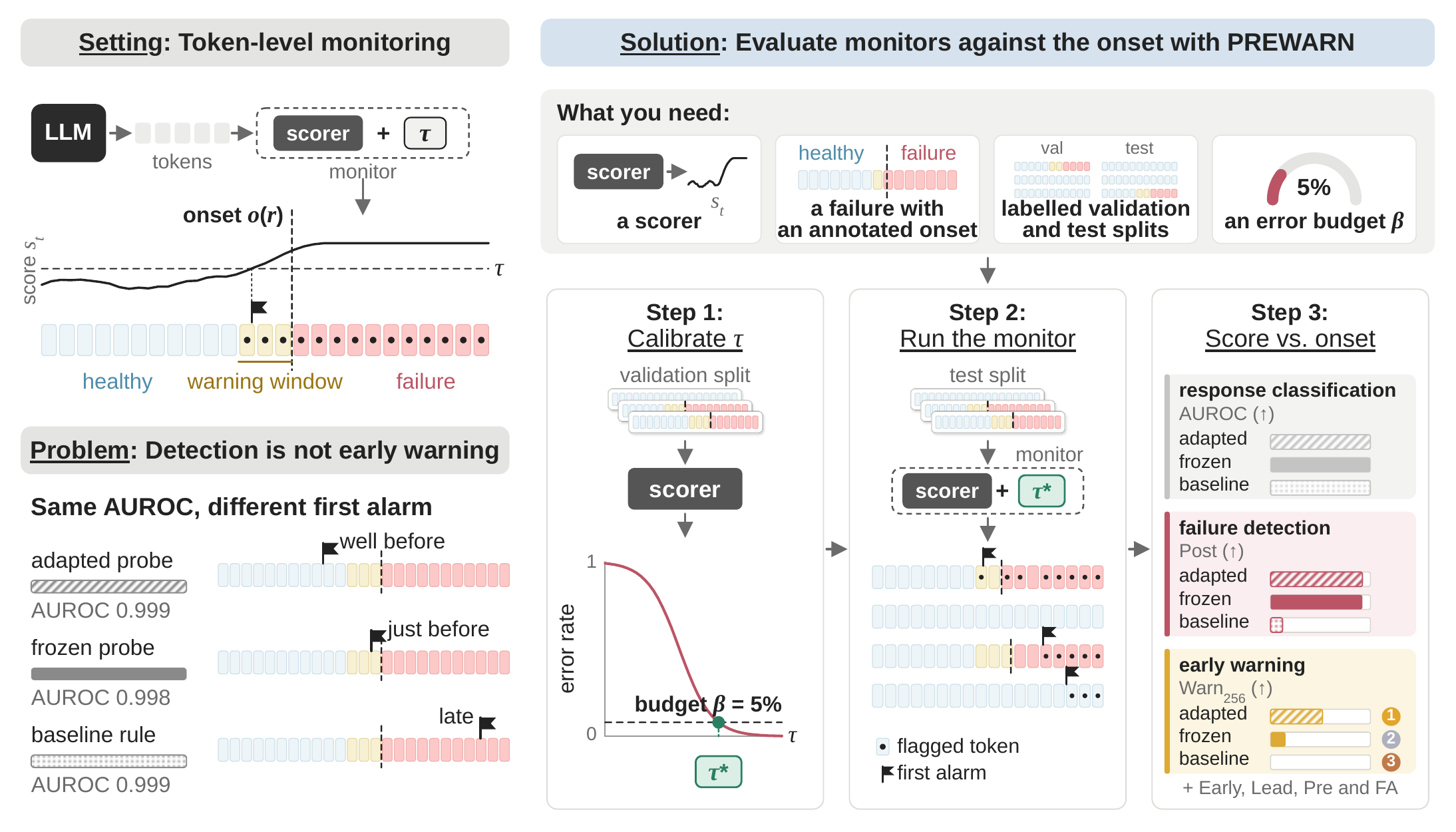}
    \caption{Overview of \method{}. Left: a monitor runs a scorer over the generated tokens and alarms where the score crosses $\tau$ (dots mark flagged tokens, the flag the first alarm); AUROC rates three such monitors as equal although their first alarms fall well before, just before and after the onset $o(r)$. Right: what \method{} needs and its three steps: calibrate the threshold on validation to the budget $\beta$, run the monitor on the test split, and score the alarms against the onset on the three tasks.}\label{fig:overview}
\end{figure}

Linear probes on model activations are a cheap way to build such monitors. They have been trained to detect deception \citep{goldowskydill2025deception}, high-stakes interactions \citep{mckenzie2025highstakes}, hallucinated entities \citep{obeso2025realtime} and repetition loops \citep{xie2025wordsalad,yu2025breaking,duan2026circular}, and are scored as classifiers of whole responses, entities, text chunks or sentences, with AUROC, accuracy, $F_1$ or recall at a fixed false-positive rate. Because every unit after the onset counts as a positive, these metrics reward flagging the failure, not flagging it early. A probe that fires only once the failure is visible in the text scores as well as one that fires before it begins (\Cref{fig:overview}, left).

To score early warning, an evaluation needs two things: an annotated onset, so that alarms before it can be scored separately from those after it, and a threshold fixed in advance at the error rate the monitor will be deployed with, so that the monitor is evaluated as it will be used. Evaluations that report timing each miss at least one of these. \citet{li2025fineharm} report how much of the response their monitor of harmful content reads before stopping it, not whether the harm had already begun. Evaluations that locate the onset either count an earlier alarm the same as one raised in the first unsafe sentence \citep{kavumba2026streamguard} or give it no credit \citep{qwen2025qwen3guard,itkin2026quickest}. \citet{huang2026prefixguard} fix the false-alarm rate, but label only whether a run failed, so they measure how early an alarm comes from the end of the run. \citet{duan2026circular} credit only alarms raised before repetition starts, but tune their threshold for detection. \Cref{app:related} covers related work in full.

We propose \method{} (\emph{pre-onset warning evaluation}), a protocol for monitors of any failure with an annotated onset. It fixes each monitor's threshold on validation data at a chosen error rate and evaluates the monitor on each of the three tasks, measuring its alarms relative to the onset (\Cref{fig:overview}, right). Which error to fix, and at what rate, depends on the failure at hand. For repetitive degeneration we fix the share of \emph{healthy} (non-degenerate) responses that raise an alarm, the \emph{false-alarm budget}. For harmful content, a missed failure can cost more than a false alarm \citep{li2025fineharm}, so one may instead fix the false-negative rate.

We apply \method{} to repetitive degeneration in \textsc{Apertus-8B-Instruct} \citep{apertus2025}, \textsc{Llama-3.1-8B-Instruct} \citep{grattafiori2024llama} and \textsc{Mistral-7B-Instruct} \citep{jiang2023mistral}, sampled stochastically without balancing the classes, and test whether response-level metrics, the standard for response classification, also identify the monitors that warn early. For repetitive degeneration they do not. Only a few percent of responses degenerate, so a monitor that never alarms is highly accurate, and since every degenerate response runs to the token limit, a rule that only checks length reaches an almost perfect AUROC, though it fires only once the budget is spent and so can never warn.

Probes that AUROC ranks as equal warn very differently under \method{}, and on \textsc{Apertus}, selecting by AUROC picks one that warns less. \emph{Frozen} probes, which read the model's own activations, raise their first alarm before the onset on most loops of \textsc{Apertus} and \textsc{Llama}, while on \textsc{Mistral} they mostly fire after the onset and work better as detectors. Low-rank adapters trained below the probe more than double how much of the window just before the onset it flags, mostly with flags long before it, but barely raise how often the first alarm precedes it. None of six untrained baselines, which read only the text or the model's confidence, warns much more often than random alarms.

Our contributions are:

\begin{itemize}

\item \method{}, a protocol that evaluates token-level monitors of any failure with an annotated onset on all three tasks, at an error rate fixed in advance (\Cref{sec:prewarn}).

\item Evidence that response-level AUROC and accuracy saturate on repetitive degeneration and only weakly predict which probes warn early (\Cref{sec:rq-metrics}).

\item A comparison with untrained baselines, in which only the probes warn before most loops on two of three models, and an analysis of what low-rank adapters add (\Cref{sec:rq-signal,sec:rq-adaptation,sec:rq-baselines}).

\item Frozen and adapted linear probes for \textsc{Apertus}, \textsc{Llama} and \textsc{Mistral}, released with thresholds calibrated at $1\%$, $5\%$ and $10\%$ at \href{https://huggingface.co/collections/marcodene/prewarn-degeneration-probes}{\nolinkurl{huggingface.co/collections/marcodene/prewarn-degeneration-probes}}.

\end{itemize} The code for training, scoring and evaluating the probes is available at \href{https://github.com/swiss-ai/degeneration-probe}{\nolinkurl{github.com/swiss-ai/degeneration-probe}}.

\section{Preliminaries} \label{sec:prelim}

\paragraph{Responses and activations.} We study a decoder-only Transformer $f$ with $L$ layers, numbered $0$ to $L-1$, and hidden size $d$. A response $r=(r_1,\ldots,r_T)$ is sampled from $f$ given a prompt, and $h_t^{(\ell)}\in\mathbb{R}^d$ is the residual-stream activation after layer $\ell$ when token $r_t$ is generated.

\paragraph{Failures and onset.} We consider failures that begin at a specific token. A response \emph{fails}, $F(r)=1$, if it contains the failure, and is \emph{healthy}, $F(r)=0$, otherwise. The \emph{onset} $o(r)$ of a failing response is the position of the first token of the failure, and is annotated together with $F(r)$. Our case study is repetitive degeneration, where a response fails if it enters a repetitive pattern that persists until the token budget is exhausted, and the onset is the first token of the first loop (\Cref{sec:setup}).

\paragraph{Scorers and monitors.} A \emph{scorer} assigns a score $s_t\in[0,1]$ to every position $t$ using only information available when $r_t$ is generated, such as the prefix or the activation $h_t^{(\ell)}$, so it can run during decoding. A \emph{monitor} runs a scorer with a threshold $\tau$: position $t$ is \emph{flagged} when $s_t\geq\tau$, and the \emph{first alarm} $\alpha_r$ of a response is its first flagged position, with $\alpha_r=\infty$ if none is flagged.

\paragraph{Monitoring tasks.} A monitor serves one of three tasks (\Cref{sec:introduction}, \Cref{fig:overview}). \emph{Response classification} asks whether $F(r)=1$ once the response is complete. \emph{Failure detection} asks whether the failure is under way at position $t$, that is, whether $t\geq o(r)$. \emph{Early warning} asks whether $t$ lies in the \emph{warning window}, the $W$ positions just before the onset, $o(r)-W\leq t<o(r)$.

\section{\texorpdfstring{\method{}}{Prewarn}} \label{sec:prewarn}

\method{} evaluates a monitor on the three tasks for any failure with an onset. It needs the token scores of a scorer on a validation and a test split, the labels of both splits, that is, which responses fail and where their onset lies, and an \emph{error budget} $\beta$, the highest rate of one kind of error the monitor may make. Which error is bounded, and at what rate, depends on the task and the application, and is left to whoever deploys the monitor. The protocol has three steps.

\paragraph{Step 1: Calibrate the threshold.} The threshold $\tau$ is set to the lowest value at which the monitor's error rate on the validation split stays within the budget, then held fixed on the test split, so that the monitor is evaluated as it will be deployed. For repetitive degeneration we choose $\beta$ to be the \emph{false-alarm budget}, the largest share of healthy responses allowed to raise an alarm. With $\mathcal{H}$ the healthy validation responses, the threshold is

\begin{equation} \tau^{*}(\beta)=\min\{\tau : |\{r\in\mathcal{H} : \alpha_r<\infty\}|\leq\beta\,|\mathcal{H}|\}. \label{eq:calibration} \end{equation}

\paragraph{Step 2: Run the monitor.} The monitor runs the scorer with $\tau$ over every test response, which gives the flags and the first alarm $\alpha_r$ of each response (\Cref{sec:prelim}). A monitor may also require several consecutive flags before it raises an alarm; \Cref{app:persistence} examines this variant.

\paragraph{Step 3: Score the alarms against the onset.} Let $\mathcal{F}$ be the failing responses of a split and $\mathcal{A}\subseteq\mathcal{F}$ those that raise an alarm. A \emph{coverage} is the share of flagged positions in a range, pooled over the responses in $\mathcal{F}$. We group the metrics by the task they measure, as in \Cref{tab:baselines}.

\begin{itemize}

\item \textbf{Response classification.} The \emph{response-level metrics} are the AUROC and the best accuracy of the response score $\max_t s_t$.

\item \textbf{Failure detection.} The \emph{detection rate} (Detection) $|\mathcal{A}|/|\mathcal{F}|$ is the share of failing responses that raise an alarm at all. \emph{Post-onset coverage} (Post) is the coverage of the positions $t\geq o(r)$.

\item \textbf{Early warning.} \emph{Warning coverage} $\mathrm{Warn}_W$ is the coverage of the warning window, with $W=256$. The \emph{early-warning rate} (Early) is the share of responses in $\mathcal{A}$ whose first alarm precedes the onset, $\alpha_r<o(r)$. The \emph{lead} (Lead) is the median of $\alpha_r-o(r)$ over $\mathcal{A}$, negative when the first alarm is a warning and positive when it is a detection. Early and Lead are computed over $\mathcal{A}$ only, so they are read together with the detection rate. Warning coverage rises when a scorer simply flags more tokens everywhere, so it is read together with pre-window coverage (below), and the early-warning rate is read against random alarms, a chance reference that flags each token with a fixed probability calibrated to the same budget. Their first alarm is equally likely to fall anywhere in a response, so their early-warning rate reflects only where the onsets lie, and a scorer warns only to the extent that it exceeds this level.

\item \textbf{Other.} \emph{Pre-window coverage} (Pre) is the coverage of the positions before the warning window, $t<o(r)-W$. Pre has no preferred direction, since a high value can reflect an early judgement that the response will fail or flags unrelated to the onset. The \emph{false-alarm rate} (FA) is the share of healthy responses of a split that raise an alarm. Calibration sets it to at most $\beta$ on validation, and on a test split it shows whether the threshold carries over.

\end{itemize} We call the detection and early-warning metrics the \emph{onset-aligned} metrics. \Cref{app:metrics} gives the full definitions.

\paragraph{Probes for early warning.} We propose linear probes on activations as scorers for early warning, with the training strategy and selection rule below, which did best among the $26$ strategies we compared on degeneration (\Cref{app:training-strategies}). A probe reads the residual stream at one layer, $s_t=\sigma\!\left(w^\top\operatorname{LN}(h_t^{(\ell)})+b\right)$, where the affine parameters of the layer norm are trained together with $w$ and $b$. A \emph{frozen} probe trains only these parameters and reads the generator's own activations as they are computed, so it needs no extra forward pass. An \emph{adapted} probe also trains rank-$16$ LoRA adapters \citep{hu2022lora} on every linear projection of the attention and MLP sublayers of layers $0$ to $\ell$, jointly with the probe, as \citet{obeso2025realtime} do for hallucination probes. Since the generator itself must stay unchanged, an adapted probe needs a second forward pass through the adapted layers.

Probes are trained with binary cross-entropy on a target that is positive from $K=256$ tokens before the onset onward, $y_t=\mathbf{1}[t\geq o(r)-K]$, which asks the probe to fire before the failure is visible in the text, on positions drawn from a window around the onset (\Cref{app:training-strategies}). Frozen and adapted probes are trained at every layer from $1$ to $L-1$, and we select the layer and checkpoint with the highest validation $\mathrm{Warn}_{256}$ at $\beta=1\%$ among those whose post-onset coverage is at least $0.3$, so that a probe is not chosen for warnings it raises without detecting the failure that follows.

\section{Experiments} \label{sec:experiments}

\subsection{Experimental setup} \label{sec:setup}

\paragraph{Models and responses.} We apply \method{} to repetitive degeneration in \textsc{Apertus-8B-Instruct} \citep{apertus2025}, \textsc{Llama-3.1-8B-Instruct} \citep{grattafiori2024llama} and \textsc{Mistral-7B-Instruct} \citep{jiang2023mistral}, all with $L=32$. For each model we sample $10$ responses to each of $3{,}630$ prompts from seven sources that span mathematics, code, instruction following and medical reasoning, with temperature $0.7$, top-$p$ $0.9$ and at most $T_{\max}=4096$ new tokens. A response is \emph{terminated} if it emits the end-of-sequence token before $T_{\max}$, and \emph{capped} otherwise. Splits are assigned per prompt. Five sources provide the training, validation and in-domain test prompts ($1{,}701$, $364$ and $365$), and two are held out entirely and used only for testing ($1{,}200$ prompts).

\paragraph{Labels.} Since every degenerate response is capped, only capped responses need a label, and since repetition can also be legitimate, as in enumerations or code, \textsc{Claude Sonnet 5} assigns it as an LLM judge \citep{zheng2023judging}. It reads the prompt and the response and quotes where the loop begins, which we locate in the tokens to obtain $o(r)$. Terminated responses are healthy, and capped responses the judge cannot rule on are left out. We then review by hand every capped response of both test splits of each model, and confirm the label and the onset, move the onset, or overturn the label. The main text reports on this \emph{human-verified test split}. The reviewers confirm $97\%$ to $99\%$ of the judge's degenerate labels and keep its onset for $74\%$ to $92\%$ of the confirmed responses (\Cref{app:data}).

\paragraph{Baselines.} Each baseline is a fixed formula with no learned weights that uses only information available when $r_t$ is generated, so it can run during generation. We group them with the probes by what they read: only the length of the response, the emitted tokens, the generator's confidence in them, or its activations. The \emph{token-cap rule} scores the last token of a capped response $1$ and every other token $0$. \emph{Windowed seq-rep-$n$} and \emph{rep}$/\ell$ measure how many of the last $S$ tokens repeat earlier ones \citep{welleck2020neural}, and the \emph{repeated-$n$-gram rule} counts how often the $n$-gram ending at $t$ has already occurred, a criterion \citet{pipis2025wait} use to label looping responses. \emph{Windowed entropy} and \emph{windowed negative log-probability} invert the generator's average uncertainty over the last $S$ tokens, since a loop is generated with high confidence, and are standard comparisons for probes that detect hallucinations \citep{manakul2023selfcheckgpt,obeso2025realtime}. \Cref{app:baselines} gives the definitions. The window $S$, the look-back $\ell$ and the $n$-gram size $n$ are chosen per model on validation with the rule that selects the probes, so each baseline is tuned and calibrated as the probes are.

\paragraph{Reported probes.} Except in \textbf{Q1}, which compares training strategies on \textsc{Apertus}, all experiments use the three models and the probes of \Cref{sec:prewarn}, with three seeds per probe. For each model we report the frozen probe at its selected layer ($15$ for \textsc{Apertus} and \textsc{Llama}, $5$ for \textsc{Mistral}), the adapted probe at the same layer, so that only the adaptation differs, and the adapted probe at its own selected layer ($30$, $6$ and $11$), which also changes how many layers are adapted. Unless stated otherwise we report $\beta=5\%$ on the human-verified in-domain test split, as the mean $\pm$ one standard deviation over seeds. \Cref{app:labels} reports the judge-labelled in-domain test split and the held-out split under both labels, and \Cref{app:budgets} reports $\beta=1\%$ and $\beta=10\%$.

We organise the experiments around four questions.

\begin{itemize}

\item[\textbf{Q1}] Can response-level metrics tell which scorers warn early?

\item[\textbf{Q2}] Do linear probes warn before the onset on every model?

\item[\textbf{Q3}] What does low-rank adaptation add to the probe?

\item[\textbf{Q4}] Do untrained baselines warn as early as the probes?

\end{itemize}

\subsection{Q1: Can response-level metrics tell which scorers warn early?} \label{sec:rq-metrics}

Since our case study is early warning of degeneration, we evaluate the probes with the early-warning metrics of \method{}. We first show why the response-level metrics of prior work cannot do this.

\paragraph{Setup.} We compare four scorers on \textsc{Apertus} at layer $15$ on the three monitoring tasks of \Cref{sec:prelim}: the token-cap rule of \Cref{sec:setup}, which fires based only on the length of the response, the adapted probe, and the two frozen probes with the highest and the lowest validation AUROC at this layer among the training strategies whose three seeds all pass the post-onset floor (\Cref{app:ranking}). We also read the frozen probe of \Cref{fig:q2} across layers.

\begin{figure}[t] \centering \includegraphics[width=\linewidth]{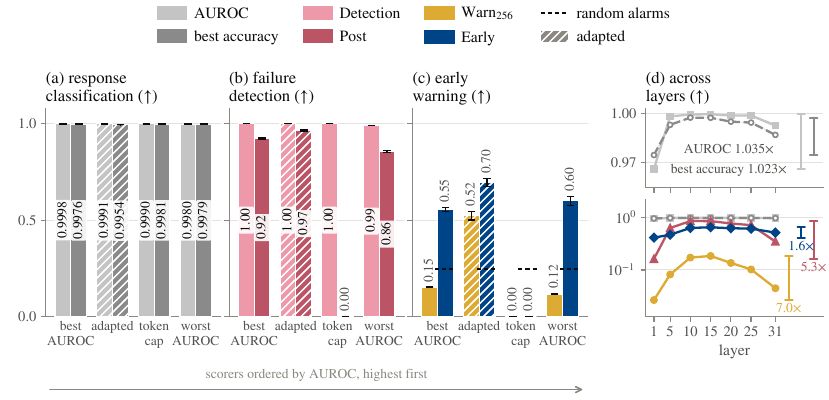} \caption{ Response-level metrics rate four scorers as equal, while only the onset-aligned metrics show how differently they perform. Scorers on \textsc{Apertus} at layer~$15$: the frozen probes with the highest and lowest validation AUROC (\Cref{app:ranking}), the adapted probe (hatched) and the token-cap rule, which flags only the last token of a capped response. Panels (a)--(c) group the metrics by task (\Cref{sec:prewarn}); dashed lines in (c) are random alarms. (d)~The frozen probe with the selected strategy across layers, with each metric's best-to-worst ratio; lower plot on a log scale. $\beta = 5\%$, human-verified in-domain test split; error bars and bands are one standard deviation over three seeds.} \label{fig:metrics} \end{figure}

\paragraph{Only the onset-aligned metrics show how differently the scorers perform, both at one layer and across layers.} On response classification the four scorers are indistinguishable, and the token-cap rule reaches the highest best accuracy (\Cref{fig:metrics}a). The detection rate, which like AUROC counts whole responses, is close to $1$ for all four (\Cref{fig:metrics}b). Differences appear only in the metrics that take the onset into account. Post-onset coverage singles out the token-cap rule, which flags almost none of the loop, while the three probes flag most of it (\Cref{fig:metrics}b). The early-warning metrics also differ between the three probes, and warning coverage, which our selection rule uses, differs most, with the adapted probe covering more than three times as much of the warning window as either frozen probe (\Cref{fig:metrics}c). Response-level metrics miss these differences because they score a response by its highest token score, wherever that token lies. For example, keeping only that token, which usually lies deep inside the loop, removes almost every early warning but leaves AUROC unchanged (\Cref{tab:positional-control} in \Cref{app:ranking}). The same pattern holds across layers (\Cref{fig:metrics}d). AUROC, which is $0.5$ for a random scorer and $1$ for a perfect one, stays above $0.96$ at every layer, and best accuracy varies by less than $3\%$, so every layer separates degenerate from healthy responses about equally well. Warning coverage is seven times higher at the best layer than at the worst, so the early signs of degeneration are much easier to read at some layers than at others.

\paragraph{AUROC ranks the probes consistently, but not by how much they warn.} In \Cref{fig:metrics}a and in the zoom of \Cref{fig:metrics}d, the response-level metrics differ little between scorers and layers, but they still show an ordering. We rank the $27$ probes of \Cref{tab:strategies} by each metric on validation and compare with their ranking on test (\Cref{tab:rank-transfer}). The ranking by AUROC reproduces itself on test (Kendall's $\tau=0.79$, against $0.53$ for best accuracy), so the ordering of AUROC is not noise, but it agrees only weakly with the test ranking by warning coverage ($\tau=0.33$). The ranking by validation warning coverage predicts it closely ($\tau=0.94$), so our selection rule uses validation warning coverage. For example, at layer $15$, selecting by validation AUROC would pick a frozen probe that covers less than a third as much of the warning window as the adapted probe.

\subsection{Q2: Do linear probes warn before the onset on every model?} \label{sec:rq-signal}

Since only the onset-aligned metrics tell scorers apart by when they fire, we use them from here on, and ask whether the probes trained for early warning (\Cref{sec:prewarn}) warn before the onset, and on which models.

\begin{figure}[t] \centering \includegraphics[width=\linewidth]{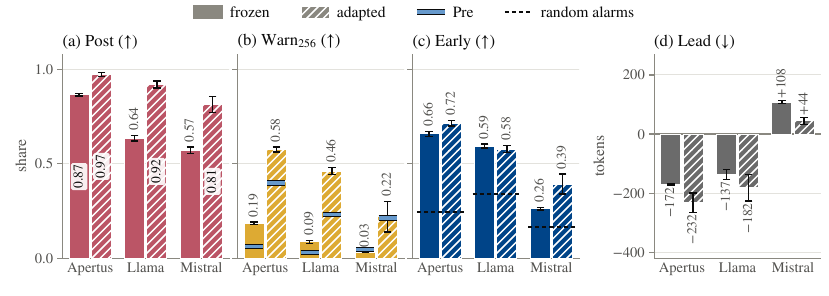} \caption{ Probes warn before most loops on \textsc{Apertus} and \textsc{Llama} but not on \textsc{Mistral}. Frozen (solid) and adapted (hatched) probe of each model at its selected layer (\Cref{tab:baselines}). (a)~Post-onset and (b)~warning coverage, with pre-window coverage (Pre) as a light-blue tick. (c)~Early-warning rate; dashed lines are random alarms. (d)~Lead, the median first alarm minus the onset in tokens; negative values are warnings. $\beta = 5\%$, human-verified in-domain test split; error bars are one standard deviation over three seeds.} \label{fig:q2} \end{figure}

\paragraph{Probes on Apertus and Llama warn before most loops.} On \textsc{Apertus} and \textsc{Llama} the early-warning rate of the frozen and the adapted probe is above one half and well above that of random alarms, and the median first alarm comes $137$ to $232$ tokens before the onset (\Cref{fig:q2}c,d). The probes also flag a larger share of the warning window than of tokens before it, with warning coverage $1.4$ to $3$ times pre-window coverage (\Cref{fig:q2}b), which suggests a signal of the coming loop.

\paragraph{Probes on Mistral fire mostly after the onset.} On \textsc{Mistral} the early-warning rate of every probe stays below one half, although above that of random alarms, and the median first alarm comes after the onset (\Cref{fig:q2}c,d). Warning coverage is close to pre-window coverage, so the probes flag the same share of tokens in the warning window as before it and find no clear signal of the approaching loop (\Cref{fig:q2}b). Their detection rate is still $0.99$ (\Cref{tab:baselines}), so on \textsc{Mistral} the probes are useful for failure detection but not for early warning. This holds for adapted probes too. Adaptation cannot add information absent from the response so far, only make it easier for a linear probe to read, so on \textsc{Mistral} the signal of the coming loop may be weak or absent until the loop begins (\Cref{sec:rq-adaptation}).

\paragraph{The error budget carries over to test, but frozen probes warn less on held-out sources.} At the $5\%$ budget, the thresholds fixed on validation give every probe a false-alarm rate of $4.5\%$ to $5.6\%$ on test and below $1\%$ on the held-out sources (\Cref{tab:heldout-human} in \Cref{app:labels}). There, the frozen probes of \textsc{Llama} and \textsc{Mistral} still raise an alarm on almost every degenerate response, but their early-warning rate drops to $0.20$ and $0.12$ and their median first alarm comes after the onset. The adapted probes degrade less and warn before almost all of the $24$ degenerate held-out responses of \textsc{Apertus}.

\subsection{Q3: What does low-rank adaptation add to the probe?} \label{sec:rq-adaptation}

The adapted probes of \textbf{Q2} flag more of the warning window than the frozen ones, so we now analyse what low-rank adaptation adds to the probe, where the added flags fall and how to interpret them.

\begin{figure}[t] \centering \includegraphics[width=\linewidth]{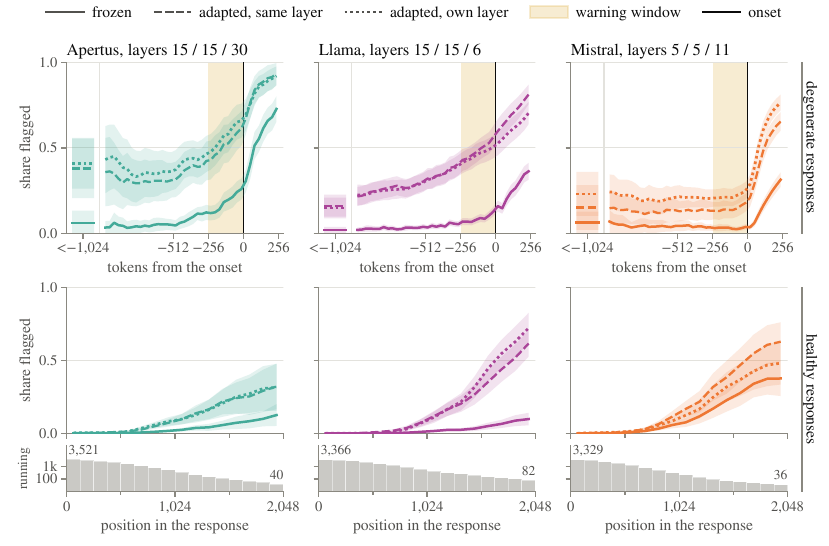} \caption{On \textsc{Apertus} and \textsc{Llama}, probes flag more tokens as the onset approaches, and adaptation adds flags long before it; on \textsc{Mistral}, flags rise only after the onset. Top: share of flagged tokens on degenerate responses by distance from the onset (bins of $32$ tokens; positions more than $1{,}024$ tokens before the onset pooled on the left). Bottom: the same on healthy responses by position (bins of $128$ tokens), where flags also rise with length; the strip below gives the number of responses still running, on a log scale. Panel titles give the layers of the three probes. $\beta = 5\%$, human-verified in-domain test split; mean over three seeds with $95\%$ intervals from leaving out one prompt at a time.} \label{fig:q4} \end{figure}

\paragraph{Adaptation more than doubles warning coverage but barely raises the early-warning rate.} We first compare the frozen probe with the adapted probe at the same layer, which differ only by the LoRA adapters. On every model adaptation more than doubles warning coverage and raises post-onset coverage, but it barely raises the early-warning rate (\Cref{tab:baselines}), so most of the early warning is already present in the frozen probe. The early-warning rate counts a response once, at its first alarm, so adding flags to a response that already has one before the onset does not raise it. Pre-window coverage rises with warning coverage, so many of the added flags fall long before the onset. Reading the adapted probe at its own layer shows the same behaviour. Since the probe stays linear, these changes come from the representation it reads. In \Cref{app:pca} we project the activations of one degenerate response per model onto their first two principal components and show that, at the frozen probe's layer, the tokens before the warning window, in it and after the onset overlap, while after adaptation they fall in largely separate regions (\Cref{fig:pca-rollouts}).

\paragraph{On Apertus and Llama, adaptation adds flags long before the onset, which on Llama may partly track response length.} \Cref{fig:q4} shows where the added flags fall. On \textsc{Apertus} and \textsc{Llama} the frozen probe flags more tokens as the onset approaches. The adapted probes show the same rise, shifted up by a level of flags already present more than a thousand tokens before the onset (top row). Much of the gain therefore comes from flags spread over the whole degenerate response, as for the long-window repetition rules of \Cref{sec:rq-baselines}. On healthy responses every probe flags more the longer the response runs (bottom row). On \textsc{Apertus} the adapted probes flag more of a degenerate response long before its onset than of a healthy response at any of the first $2{,}048$ positions, which suggests a judgement, made long before the onset, that the response will degenerate. On \textsc{Llama} healthy responses reach the same level once they run long enough, so these flags may partly reflect how long the response has run. On both models the onset typically comes after a healthy response has ended, so position alone carries part of the label, and the probes may partly rely on it.

\paragraph{On Mistral, adaptation raises flags evenly before the onset, not toward it.} On \textsc{Mistral} all three probes flag every position before the onset at about the same rate and rise only after it. Adaptation raises this rate without adding a rise toward the onset, which keeps warning coverage close to pre-window coverage (\Cref{fig:q2}b).

\subsection{Q4: Do untrained baselines warn as early as the probes?}
\label{sec:rq-baselines}

\begin{table}[t]
  \centering
  \caption{Probes against the baselines
on the human-verified in-domain test split at $\beta = 5\%$. Scorers are
grouped, from top to bottom, by what they read: nothing (random alarms, the
chance reference), only the length of the response, the emitted tokens, the
generator's confidence in them, or the activations. Each baseline is read
at the setting chosen on validation (\Cref{app:baselines}). Probes are
labelled with the layer they read; adapted probes also train rank-$16$ LoRA
adapters in the layers up to it. Metrics as defined in \Cref{sec:prewarn};
Lead is in tokens, negative when the first alarm precedes the onset. $n$:
number of degenerate responses. $\dagger$: raises an alarm on fewer than $90\%$ of degenerate
responses, so its Early and Lead describe only part of them and are left
out of the comparison. FA and Pre are not ranked. Best value in each other
column and model in bold, values within one standard deviation of it
underlined. Mean
$\pm$ one standard deviation over three seeds for probes; the baselines are
deterministic, and random alarms are given in closed form.}
  \label{tab:baselines}
  \footnotesize
  \setlength{\tabcolsep}{3pt}
  \resizebox{\linewidth}{!}{%
  \begin{tabular}{@{}lrrrrrrrr@{}}
    \toprule
    & \shortstack{Response\\classification} & \multicolumn{2}{c}{Failure detection} & \multicolumn{3}{c}{Early warning} & & \\
    \cmidrule(lr){2-2}\cmidrule(lr){3-4}\cmidrule(lr){5-7}
    Scorer & AUROC $\uparrow$ & Detection $\uparrow$ & Post $\uparrow$ & $\mathrm{Warn}_{256}$ $\uparrow$ & Early $\uparrow$ & Lead $\downarrow$ & FA (\%) & Pre \\
    \midrule
    \multicolumn{9}{@{}l}{\textsc{Apertus} ($n=116$)} \\
    \quad random alarms & $0.8901$ & $0.32$ & $1\times10^{-4}$ & $1\times10^{-4}$ & $0.246$$^\dagger$ & $+985$$^\dagger$ & $5.0$ & $1\times10^{-4}$ \\
    \quad token-cap rule & $0.9990$ & $1.00$ & $3\times10^{-4}$ & $0.000$ & $0.000$ & $+3276$ & $0.2$ & $0.000$ \\
    \addlinespace[2pt]
    \quad repeated-$n$-gram rule & $0.9972$ & $0.99$ & $0.779$ & $0.004$ & $0.139$ & $+187$ & $3.2$ & $0.003$ \\
    \quad windowed seq-rep-$n$ & $0.9815$ & $0.93$ & $0.876$ & $0.207$ & $0.296$ & $+148$ & $5.2$ & $0.128$ \\
    \quad rep$/\ell$ & $0.9983$ & $1.00$ & $0.912$ & $0.228$ & $0.267$ & $+230$ & $4.8$ & $0.150$ \\
    \addlinespace[2pt]
    \quad windowed entropy & $0.9971$ & $1.00$ & $0.787$ & $0.023$ & $0.112$ & $+536$ & $6.2$ & $0.054$ \\
    \quad \begin{tabular}[t]{@{}l@{}}windowed negative\\ log-probability\end{tabular} & $0.9964$ & $1.00$ & $0.808$ & $0.021$ & $0.129$ & $+400$ & $5.2$ & $0.053$ \\
    \addlinespace[2pt]
    \quad frozen probe, L15 & $\mathbf{0.9994\sd{0.0000}}$ & $1.00\sd{0.00}$ & $0.865\sd{0.007}$ & $0.186\sd{0.007}$ & $0.658\sd{0.013}$ & $-172\sd{3}$ & $5.6\sd{0.3}$ & $0.063\sd{0.005}$ \\
    \quad adapted probe, L15 & $0.9991\sd{0.0001}$ & $1.00\sd{0.00}$ & \underline{$0.966\sd{0.002}$} & $0.523\sd{0.022}$ & $0.695\sd{0.020}$ & \underline{$-214\sd{37}$} & $4.8\sd{0.6}$ & $0.351\sd{0.012}$ \\
    \quad adapted probe, L30 & \underline{$0.9982\sd{0.0012}$} & $1.00\sd{0.00}$ & $\mathbf{0.973\sd{0.012}}$ & $\mathbf{0.577\sd{0.013}}$ & $\mathbf{0.716\sd{0.015}}$ & $\mathbf{-232\sd{33}}$ & $4.6\sd{0.1}$ & $0.399\sd{0.020}$ \\
    \midrule
    \multicolumn{9}{@{}l}{\textsc{Llama} ($n=284$)} \\
    \quad random alarms & $0.8810$ & $0.29$ & $9\times10^{-5}$ & $9\times10^{-5}$ & $0.340$$^\dagger$ & $+632$$^\dagger$ & $4.9$ & $9\times10^{-5}$ \\
    \quad token-cap rule & $\mathbf{0.9988}$ & $1.00$ & $4\times10^{-4}$ & $0.000$ & $0.000$ & $+2958$ & $0.2$ & $0.000$ \\
    \addlinespace[2pt]
    \quad repeated-$n$-gram rule & $0.8987$ & $0.81$ & $0.608$ & $0.056$ & $0.297$$^\dagger$ & $+156$$^\dagger$ & $3.3$ & $0.015$ \\
    \quad windowed seq-rep-$n$ & $0.9953$ & $0.99$ & $0.872$ & $0.307$ & $0.358$ & $+231$ & $5.4$ & $0.185$ \\
    \quad rep$/\ell$ & $0.9981$ & $1.00$ & $0.867$ & $0.315$ & $0.359$ & $+262$ & $4.2$ & $0.193$ \\
    \addlinespace[2pt]
    \quad windowed entropy & $0.9655$ & $0.84$ & $0.364$ & $0.026$ & $0.118$$^\dagger$ & $+964$$^\dagger$ & $5.4$ & $0.034$ \\
    \quad \begin{tabular}[t]{@{}l@{}}windowed negative\\ log-probability\end{tabular} & $0.9685$ & $0.94$ & $0.462$ & $0.018$ & $0.097$ & $+945$ & $4.8$ & $0.029$ \\
    \addlinespace[2pt]
    \quad frozen probe, L15 & $0.9986\sd{0.0001}$ & $1.00\sd{0.00}$ & $0.635\sd{0.014}$ & $0.087\sd{0.008}$ & $0.594\sd{0.010}$ & $-137\sd{18}$ & $5.4\sd{0.1}$ & $0.029\sd{0.004}$ \\
    \quad adapted probe, L15 & \underline{$0.9980\sd{0.0008}$} & $1.00\sd{0.00}$ & $\mathbf{0.945\sd{0.008}}$ & $\mathbf{0.488\sd{0.009}}$ & $\mathbf{0.655\sd{0.020}}$ & $\mathbf{-236\sd{6}}$ & $5.5\sd{0.2}$ & $0.240\sd{0.007}$ \\
    \quad adapted probe, L6 & $0.9948\sd{0.0013}$ & $1.00\sd{0.00}$ & $0.920\sd{0.019}$ & $0.462\sd{0.018}$ & $0.578\sd{0.019}$ & $-182\sd{45}$ & $5.3\sd{0.2}$ & $0.231\sd{0.002}$ \\
    \midrule
    \multicolumn{9}{@{}l}{\textsc{Mistral} ($n=315$)} \\
    \quad random alarms & $0.9026$ & $0.35$ & $1\times10^{-4}$ & $1\times10^{-4}$ & $0.167$$^\dagger$ & $+1254$$^\dagger$ & $4.8$ & $1\times10^{-4}$ \\
    \quad token-cap rule & $\mathbf{0.9995}$ & $1.00$ & $3\times10^{-4}$ & $0.000$ & $0.000$ & $+3688$ & $0.1$ & $0.000$ \\
    \addlinespace[2pt]
    \quad repeated-$n$-gram rule & $0.9539$ & $0.91$ & $0.757$ & $0.008$ & $0.066$ & $+156$ & $1.8$ & $0.007$ \\
    \quad windowed seq-rep-$n$ & $0.9981$ & $1.00$ & $\mathbf{0.935}$ & \underline{$0.186$} & $0.203$ & $+192$ & $5.6$ & $0.186$ \\
    \quad rep$/\ell$ & $0.9974$ & $1.00$ & $0.916$ & \underline{$0.162$} & $0.200$ & $+301$ & $4.5$ & $0.187$ \\
    \addlinespace[2pt]
    \quad windowed entropy & $0.9928$ & $0.97$ & $0.820$ & $0.033$ & $0.075$ & $+496$ & $4.6$ & $0.050$ \\
    \quad \begin{tabular}[t]{@{}l@{}}windowed negative\\ log-probability\end{tabular} & $0.9922$ & $0.98$ & $0.816$ & $0.017$ & $0.052$ & $+347$ & $3.3$ & $0.035$ \\
    \addlinespace[2pt]
    \quad frozen probe, L5 & $0.9916\sd{0.0004}$ & $0.99\sd{0.00}$ & $0.572\sd{0.018}$ & $0.031\sd{0.002}$ & $0.262\sd{0.007}$ & $+108\sd{5}$ & $5.2\sd{0.1}$ & $0.047\sd{0.002}$ \\
    \quad adapted probe, L5 & $0.9942\sd{0.0018}$ & $0.99\sd{0.01}$ & $0.862\sd{0.054}$ & \underline{$0.147\sd{0.017}$} & $0.309\sd{0.011}$ & $+76\sd{3}$ & $5.2\sd{0.5}$ & $0.137\sd{0.012}$ \\
    \quad adapted probe, L11 & $0.9952\sd{0.0006}$ & $0.99\sd{0.00}$ & $0.813\sd{0.042}$ & $\mathbf{0.220\sd{0.081}}$ & $\mathbf{0.393\sd{0.054}}$ & $\mathbf{+44\sd{12}}$ & $4.5\sd{0.7}$ & $0.213\sd{0.073}$ \\
    \bottomrule
  \end{tabular}%
  }
\end{table}

Finally, we ask whether the warning of the probes could also come from the baselines of \Cref{sec:setup}, which need no training. \Cref{tab:baselines} compares them with the probes on the three tasks.

\paragraph{Every scorer has a high AUROC, and the token-cap rule the highest on two models.} As in \textbf{Q1}, response classification tells the scorers apart least. Every scorer but random alarms and the repeated-$n$-gram rule on \textsc{Llama} has an AUROC above $0.95$, and the token-cap rule has the highest AUROC on \textsc{Llama} and \textsc{Mistral}, although its only flag falls on the last token of a capped response, after the loop has filled the budget. On \textsc{Apertus} the adapted probe at layer $30$, which covers the most of the warning window, has a lower AUROC than both the token-cap rule and rep$/\ell$. Ranking by AUROC would again not select the scorers that warn.

\paragraph{The adapted probes flag most of the loop, but on Mistral the long-window repetition rules flag more.} Every scorer raises an alarm on at least $81\%$ of the degenerate responses, and most on nearly all of them, so the ranking on failure detection comes down to post-onset coverage. The adapted probes flag the largest share of the loop on \textsc{Apertus} and \textsc{Llama}, while on \textsc{Mistral} windowed seq-rep-$n$ and rep$/\ell$ flag more. The text baselines measure repetition in the emitted tokens, which appears only once the loop has completed its first cycle, and the confidence baselines measure how confidently the generator predicts the next token, which rises only once tokens repeat the previous cycle. Both appear after the onset, so these baselines can detect the loop but not warn of it.

\paragraph{The probes warn earliest on every model, and no baseline warns much more often than random alarms.} On every model the probes have the highest early-warning rate and the earliest median first alarm, and on \textsc{Apertus} and \textsc{Llama} they are the only scorers with an early-warning rate above one half. No baseline warns much more often than random alarms, and the token-cap, repeated-$n$-gram and confidence rules warn less often. Windowed seq-rep-$n$ and rep$/\ell$ cover more of the warning window than the frozen probe on every model, but warning coverage is not read alone (\Cref{sec:prewarn}). They flag much of the text before the warning window too (Pre), and their first alarm comes well after the onset, so they do not pick up the coming loop but flag much of the response. Both select windows of $1{,}024$ to $4{,}096$ tokens on validation (\Cref{tab:sweep-seqrep,tab:sweep-repl}), so their score measures how repetitive the response has been so far, and a response that repeats itself from the start is flagged throughout, warning window included. The adapted probe at its own layer still flags $2.5$ and $1.5$ times as much of the warning window as the best baseline on \textsc{Apertus} and \textsc{Llama} and is within one standard deviation of windowed seq-rep-$n$ on \textsc{Mistral}.

\section{Conclusion} \label{sec:conclusion}

We asked how a token-level monitor of a language model failure should be evaluated, and whether linear probes on activations can warn of repetitive degeneration before it begins. Response-level metrics score whether a response fails, not when a monitor fires, and for repetitive degeneration they rate every scorer as near-perfect, including a length-only rule. \method{} instead calibrates the threshold on validation data at whichever error rate the task calls for, runs the monitor with that threshold, and scores every alarm against the annotated onset on all three tasks (\Cref{sec:prewarn}). It needs only token scores, annotated onsets and an error budget, so it applies to any failure with an onset, such as harmful content \citep{li2025fineharm} or hallucinated entities \citep{obeso2025realtime}, which we leave to future work. Under \method{}, probes that response-level metrics rate as equal differ widely in how much they warn (\Cref{sec:rq-metrics}). Probes on \textsc{Apertus} and \textsc{Llama} fire before the onset on most of the loops they flag, while on \textsc{Mistral} their median alarm comes after the onset (\Cref{sec:rq-signal}). Low-rank adaptation below the probe more than doubles warning coverage on every model, mostly with flags long before the onset, while the early-warning rate rises little (\Cref{sec:rq-adaptation}). No untrained baseline warns much more often than random alarms (\Cref{sec:rq-baselines}).

The two kinds of probe suit different deployments. A frozen probe reads the generator's own activations at no extra cost and already provides most of the early warning. An adapted probe flags much more of the warning window and of the loop, at the cost of a second forward pass through the adapted layers. On a model like \textsc{Mistral}, where neither warns before most loops, a probe is better used as a detector. Because the probes we release come with thresholds calibrated at $1\%$, $5\%$ and $10\%$, they can directly trigger an intervention, such as stopping, resampling or steering the generation before the loop begins, a direction we have not yet explored.

\section{Limitations} \label{sec:limitations}

First, the lead times depend on the onset convention. We place the onset at the first token of the first loop, before the repetition is visible in the text, and a later convention would make every alarm come earlier relative to the onset. Second, under our definition of degeneration, a response that loops and then emits the end-of-sequence token counts as healthy, so our false-alarm rates are conservative. Third, some prompts call for repetition that is requested or informative, such as explicit requests to repeat, enumerations, brute-force search or code. These responses are labelled healthy, but we do not test whether the probes can tell such repetition apart from degeneration. Fourth, on \textsc{Apertus} and \textsc{Llama} the onset usually comes after a healthy response has ended, so part of what the probes flag may reflect how long the response has run (\Cref{sec:rq-adaptation}). Fifth, the training strategy and selection rule of \Cref{sec:prewarn} were chosen on degeneration alone. Another failure may call for a different strategy, or for scorers other than linear probes.

\begin{ack}
This research was supported by the ETH AI Center through an ETH AI Center postdoctoral fellowship to Anna Hedstr\"{o}m. This work was supported as part of the Swiss AI Initiative by compute grant infra01 from the Swiss National Supercomputing Centre (CSCS) on Alps.
\end{ack}

\bibliographystyle{plainnat}
\bibliography{references}

\newpage
\appendix

\section{Appendix}
\label{app:appendix}

\FloatBarrier
\subsection{Related work in full}
\label{app:related}

\paragraph{Repetitive degeneration.}
\citet{holtzman2020curious} show that repetition, a long-standing failure of
neural text generation, is most frequent under
likelihood-maximising decoding, with $73.66\%$ of greedy continuations ending
in repetition, and \citet{xu2022learning} show that the more often a sentence
already appears in the context, the more likely the model is to generate it
again. \citet{welleck2020neural} trace repetition to the training objective and
define the lexical measures \texttt{seq-rep-n} and rep$/\ell$. The failure has
re-emerged in reasoning models, where a loop consumes the remaining decoding
budget \citep{xie2025wordsalad,duan2026circular}, and \citet{pipis2025wait}
find that raising the temperature reduces looping without removing its cause.

\paragraph{Detectors of degeneration.}
\citet{xie2025wordsalad} classify
each chunk of a reasoning trace from the final layer, reaching
$89.8$--$93.5\%$ accuracy on R1-Distill-Qwen-7B at temperature $0$, and stop the
generation after two consecutive flagged long chunks (or five short ones). \citet{yu2025breaking} detect recurrent
generation from the similarity of MLP activation patterns across all layers,
on prompts found by a search for inputs that trigger loops.
\citet{duan2026circular} classify sentences from their last-layer states and
accumulate the scores with a cumulative-sum rule. They count a detection only
if it precedes a rule-based start of repetition, and report early-detection
rates of $0.64$--$0.76$ at false-positive rates of $0.24$--$0.34$ and lead times
of $1306$--$1872$ tokens, with greedy decoding, balanced test sets and
thresholds tuned on held-out data. \citet{zhang2026radar} read attention
statistics and report that, averaged over responses, their score crosses
moderate thresholds before the recorded onset, on loops induced by adversarial
or missing-premise prompts, and \citet{oladri2026token} probe residual states to
predict that a reasoning model will exhaust its budget, with an AUROC of $0.61$
on one model. These detectors read the model at a fixed place, the final layer or all
layers together, although linear features are often most readable in the middle
of the network \citep{skean2025layer}. We therefore read our probes at every
layer. None of these works evaluates token-level alarms against an annotated
onset under a fixed false-alarm budget, and none studies naturally occurring
loops in non-reasoning instruction-tuned models under stochastic sampling.

\paragraph{Monitors for other failures with an onset.} Linear probes on activations classify deceptive responses \citep{goldowskydill2025deception}, reporting recall at a fixed false-positive rate on control data, classify high-stakes interactions \citep{mckenzie2025highstakes}, and flag hallucinated entities while they are generated \citep{obeso2025realtime}. Probes on prompt activations can also predict response-level behaviours, such as format failures or jailbreak compliance, before generation starts \citep{ashok2025predict}, but each prompt then gets a single prediction, so timing within the response is not evaluated. Streaming monitors of harmful content read each prefix and can halt the output \citep{li2025fineharm,qwen2025qwen3guard}, and \citet{kavumba2026streamguard} forecast the harm of likely continuations from a prefix. \citet{kavumba2026streamguard} count an alarm as timely if it fires by the end of the first unsafe sentence, so an alarm raised before the unsafe content begins and one raised partway through that sentence count the same. \citet{qwen2025qwen3guard} count an alarm as a hit only if it falls inside that sentence and report earlier alarms separately, without credit. StreamGuard reports false positives at a default threshold, and the latency evaluation of Qwen3Guard contains only unsafe responses. \citet{singteam2026singprobe} annotate where the unsafe part starts but score guards only with response- and token-level AUROC. \citet{itkin2026quickest} measures the delay of hallucination detectors after the annotated onset at a fixed per-token false-alarm rate, and charges an alarm raised before the onset as a miss. PreAct-Bench \citep{xu2026preact} asks LLM judges, guard models and a linear probe to predict, from an agent trajectory cut before the first overtly unethical step, whether the agent will act unethically, and scores them by $F_1$ on balanced pairs without a false-alarm budget. Concurrent work fixes the false-alarm rate of language-model monitors on runs without the failure \citep{schirmer2026online,dubey2026realtime,gao2026probe,huang2026prefixguard} and argues that AUROC does not establish a valid alarm policy \citep{gao2026probe}, but none credits alarms raised before an annotated onset. \citet{huang2026prefixguard} warn of failures in agent runs at a budget set on successful runs, but take their labels from the final outcome only and measure lead time from the end of the run.

\paragraph{Evaluating early-warning systems.}
Seizure prediction evaluates
sensitivity together with the false prediction rate, for a given prediction
horizon and occurrence period \citep{winterhalder2003seizure}, and clinical
systems count an event as caught if an alarm fires within a window before it,
reporting precision and alarms per patient-hour at a fixed event recall
\citep{hyland2020circulatory}. Security work compares attacks by their
true-positive rate at a low false-positive rate, which AUROC hides
\citep{carlini2022membership}. \citet{kamran2024sepsis} report that the Epic
Sepsis Model reaches an AUROC of $0.87$ over all its predictions, but $0.62$ on
predictions made before sepsis criteria are met and $0.47$ on predictions made
before clinical recognition (the first culture, lactate, antibiotic or fluid
order), suggesting that it partly reacts to clinicians' suspicion. A
token-level monitor scored with pooled metrics can look good in the same way,
by firing once the failure is already visible. \citet{yang2026prediction} argue
that clinical early-warning models should be judged against an independently
adjudicated onset, sorting each alarm as before, around or after it, and
reporting false alarms at the chosen threshold. For dialogs and agent
trajectories, \citet{baidya2026sparse} compare failure alerts by their
trade-off between accuracy and earliness, measured as the trigger turn over the
trajectory length.

\paragraph{Acting on an alarm and adapting the probe.}
Activation steering can be applied only when a condition read from the activations holds
\citep{lee2025cast} or when a calibrated estimate says it will reduce the error
\citep{hedstrom2025mera}, and a model can be trained to discard an unsafe
partial response and regenerate it \citep{zhang2025backtracking}. For
degeneration, \citet{xie2025wordsalad} cut the trace and regenerate once a loop
is detected. These interventions act on a failure that is already under way or
on a prediction made from the prompt, and acting on a warning raised before the
onset of degeneration is left to future work. To make the warning easier to
read, low-rank adapters \citep{hu2022lora} can be trained in the layers below a
linear probe jointly with it, which raises the AUROC of hallucination probes
\citep{obeso2025realtime}. Fine-tuning tends to strengthen mechanisms the model
already has \citep{prakash2024finetuning}, but it can also change what a probe
reads, to the point of hiding a harmful state from it
\citep{bailey2026obfuscated}.

\FloatBarrier
\subsection{Data and labels}
\label{app:data}

\paragraph{Prompts and responses.}
The prompts come from seven sources: mathematics (\href{https://huggingface.co/datasets/zwhe99/DeepMath-103K}
{\texttt{deepmath\_103k}}
\citep{he2025deepmath}, \href{https://huggingface.co/datasets/AI-MO/NuminaMath-1.5}
{\texttt{numinamath\_1\_5}} and \href{https://huggingface.co/datasets/MathArena/aime_2025}
{\texttt{aime\_2025}}), instruction following
(\href{https://huggingface.co/datasets/allenai/IF_sft_data_verified}
{\texttt{if\_sft\_data\_verified}}), reasoning traces from the
\href{https://huggingface.co/datasets/nvidia/Llama-Nemotron-Post-Training-Dataset}{\texttt{llama\_nemotron}}
post-training data \citep{bercovich2025nemotron}, competitive programming
(\href{https://huggingface.co/datasets/open-r1/codeforces}{\texttt{codeforces}}) and medical reasoning (\href{https://huggingface.co/datasets/FreedomIntelligence/medical-o1-verifiable-problem}{\texttt{medical\_o1}}
\citep{chen2024huatuogpt}). Each prompt is answered $10$ times by each model,
with temperature $0.7$, top-$p$ $0.9$ and at most $4{,}096$ new tokens. Splits
are assigned per prompt, so all responses to a prompt fall in the same split,
and the two held-out sources appear only in the held-out test split.
\Cref{tab:corpus-splits} gives the sizes.

\begin{table}[!htbp]
  \centering
  \caption{Dataset size per split: prompts, and degenerate responses per
model with their share of evaluated responses ($10$ responses per prompt and
model, minus those the judge could not label). \texttt{codeforces} and
\texttt{medical\_o1} appear only in the held-out split. Test rows use the
human-verified labels, train and validation rows the judge's.}
  \label{tab:corpus-splits}
  \small
  \begin{tabular}{lrrrr}
    \toprule
    & & \multicolumn{3}{c}{Degenerate responses} \\
    \cmidrule(lr){3-5}
    Split & Prompts & \textsc{Apertus} & \textsc{Llama} & \textsc{Mistral} \\
    \midrule
    Train & 1{,}701 & 567 (3.3\%) & 1{,}411 (8.3\%) & 1{,}668 (9.8\%) \\
    Validation & 364 & 108 (3.0\%) & 272 (7.5\%) & 389 (10.7\%) \\
    Test, in-domain & 365 & 116 (3.2\%) & 284 (7.8\%) & 315 (8.6\%) \\
    Test, held-out & 1{,}200 & 24 (0.2\%) & 133 (1.1\%) & 129 (1.1\%) \\
    \bottomrule
  \end{tabular}
\end{table}

\paragraph{Judge.}
\textsc{Claude Sonnet 5} labels every capped response. It receives the prompt
and the response, decides whether the response degenerates, and if it does
returns a short verbatim quotation of where the repetitive pattern begins.
The quotation marks the first occurrence of the pattern, not a later repeat,
even when it is not yet clear at that point that the pattern will repeat. We
locate the quotation in the tokenised response, and its first token is the
onset $o(r)$. Terminated responses are treated as healthy without judging. A
response that loops and then emits the end-of-sequence token therefore counts
as healthy, and an alarm on it is charged to the false-alarm budget, which
makes the budget conservative. Responses whose quotation cannot be located, or on which
the judge fails, are left out of every split. Capped responses the judge rules
healthy are healthy responses like the terminated ones.

\paragraph{Judge system prompt.}
The judge receives only the original prompt and generated completion; it does
not observe any of the heuristic repetition scores used elsewhere in our
analysis. We use the following system prompt:

\begin{quote}
\small
\ttfamily
\raggedright

You are judging whether a language model's completion degenerated into
repetitive, low-information text (looping on a phrase, sentence, or pattern
without making progress), as opposed to repetition that is a normal, expected
part of the task. You will see both the prompt and the completion -- read the
prompt first, since it changes what counts as degeneration.

\medskip

Two things that look repetitive but are NOT degeneration:

\medskip

1. Genuine incremental work. Brute-force search, case enumeration, or a
step-by-step derivation often produces a fixed template with a changing value
each time (e.g. ``N = 41: ...'' then ``N = 42: ...'' then ``N = 43: ...'').
This is fine as long as each occurrence is doing new work -- checking a new
candidate, advancing to the next step, covering a new case. A useful test: if
you deleted every occurrence but the first, would the completion lose
information needed to solve the task? If yes, it's legitimate enumeration, not
degeneration. Only flag it if the ``progress'' is illusory -- e.g. it starts
re-checking values it already covered, drifts into a template that stops
advancing, or cycles without ever converging on an answer.

\medskip

2. Instructed repetition. If the prompt itself asks for repeated output (e.g.
``repeat X ten times'', ``write a line for each of the following ...''),
producing that repetition is correct instruction-following, not degeneration --
unless the completion overruns what was asked (keeps going well past the
requested count) or the repetition itself becomes corrupted or loops in a way
the prompt didn't ask for.

\medskip

Completions are all hard-capped at the same fixed max-token budget. A
completion that ends abruptly mid-sentence, mid-word, or mid-token is expected
simply because it hit that cap -- this is not, by itself, evidence of
degeneration one way or the other. Judge based on the quality and content of
the text alone, not on whether or how it was cut off.

\end{quote}

\paragraph{Human verification.}
Every capped response of the two test splits that the judge rules on is read
by one of four reviewers, $149$ for \textsc{Apertus}, $425$ for \textsc{Llama} and $450$ for
\textsc{Mistral}. The reviewer sees the prompt, the full response and the
judge's verdict, onset and rationale, and either confirms the verdict and the
onset, moves the onset, or overturns the verdict. The human-verified test split
takes the reviewer's verdict and onset for every response. The reviewers
confirm $139$ of the $143$ degenerate verdicts on \textsc{Apertus} ($97\%$),
$417$ of $424$ on \textsc{Llama} ($98\%$) and $444$ of $450$ on
\textsc{Mistral} ($99\%$), and of the seven capped responses the judge rules
healthy they mark one as degenerate. Their notes describe most of the $17$
overturned verdicts as enumerations or brute-force searches that still do work
for the task, the kind of repetition the judge prompt asks it not to flag.

\begin{figure}[h]
  \centering
  \includegraphics[width=\linewidth]{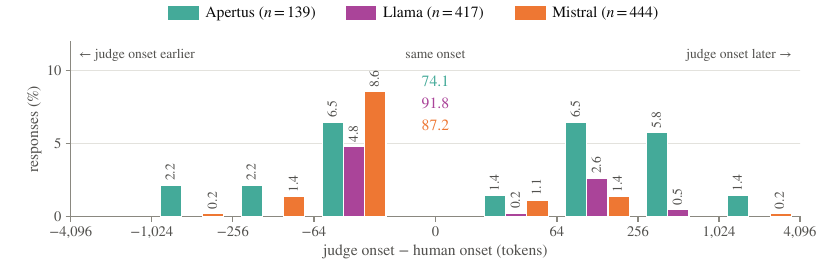}
  \caption{Signed distance between the judge's onset and the reviewer's onset
  on the capped test responses that both call degenerate, pooled over the two
  test splits, in log-spaced bins. Negative values mean the judge places the
  onset earlier than the reviewer. Every number is a percentage of the model's
  responses. The bars give the responses whose onset moved and the middle
  column those whose onset the reviewer kept, so the numbers of one model add
  up to $100$.}
  \label{fig:judge-onset}
\end{figure}

When both call a response degenerate, the reviewer keeps the judge's onset in
$74\%$ of the \textsc{Apertus}, $92\%$ of the \textsc{Llama} and $87\%$ of the
\textsc{Mistral} responses (\Cref{fig:judge-onset}). On \textsc{Llama} and
\textsc{Mistral}, $97\%$ of the judge's onsets lie within $64$ tokens of the
reviewer's, and only two responses per model are off by $256$ tokens or more,
the width of the warning window. \textsc{Apertus} has a longer tail. Thirteen
of its $139$ onsets are off by $256$ tokens or more, and on ten of these the
judge places the onset $380$ to $1{,}250$ tokens after the reviewer does. A late
onset moves the warning window into degenerate text, where an alarm counts as a
warning although it comes after the loop has begun. Consistent with this, the
human-verified labels lower warning coverage on \textsc{Apertus} by up to
$0.05$ on the in-domain split and up to $0.12$ on the held-out split at
$\beta=5\%$, while on \textsc{Llama} and \textsc{Mistral} no scorer moves by
more than $0.011$ (\Cref{app:labels}). Each response is read by a single
reviewer, so we do not measure how often two reviewers agree.

\FloatBarrier
\subsection{Metric definitions} \label{app:metrics}

Let $\mathcal{F}$ be the failing responses of a split, $T_r$ the length of response $r$, and $f_t=\mathbf{1}[s_t\geq\tau^{*}(\beta)]$ the flag of token $t$. Coverage is pooled over tokens: \begin{align} \mathrm{Post} &= \frac{\sum_{r\in\mathcal{F}}\sum_{t=o(r)}^{T_r} f_t}{\sum_{r\in\mathcal{F}} (T_r-o(r)+1)}, & \mathrm{Warn}_W &= \frac{\sum_{r\in\mathcal{F}}\sum_{t=\max(1,o(r)-W)}^{o(r)-1} f_t}{\sum_{r\in\mathcal{F}} \min(W,o(r)-1)}. \end{align}
\emph{Pre-window coverage} (Pre) is defined like $\mathrm{Warn}_W$ over the
positions $1\leq t<o(r)-W$ before the warning window. With $\mathcal{A}$ the failing
responses whose first alarm $\alpha_r$ (\Cref{sec:prelim}) is finite, the
detection rate is $|\mathcal{A}|/|\mathcal{F}|$, the early-warning rate is
$|\{r\in\mathcal{A}:\alpha_r<o(r)\}|/|\mathcal{A}|$, and the lead is the
median of $\alpha_r-o(r)$ over $\mathcal{A}$. With $\mathcal{H}'$ the healthy
responses of the split, the false-alarm rate is
$|\{r\in\mathcal{H}':\alpha_r<\infty\}|/|\mathcal{H}'|$. The response-level metrics use
the response score $\max_t s_t$. AUROC is the area under its ROC curve, and best
accuracy is the highest accuracy any threshold on it reaches, including the
threshold that flags nothing.

\paragraph{Persistence.} The main text raises an alarm at the first flagged
token. With a \emph{persistence} of $m$ tokens, the first alarm is instead the
first position that completes a run of $m$ consecutive flagged tokens,
\begin{equation}
  \alpha_r=\min\{t : s_{t'}\geq\tau\ \text{for all}\ t'\in(t-m,t]\},
  \label{eq:alarm}
\end{equation}
with $\alpha_r=\infty$ if no such run exists, and the threshold is calibrated
on this $\alpha_r$ (\Cref{eq:calibration}). With $m=1$ it is the first flagged
token. \Cref{app:persistence} compares $m\in\{1,16,64\}$.

\FloatBarrier
\subsection{Training strategies} \label{app:training-strategies}

All training strategies train the probe of \Cref{sec:prewarn} with binary cross-entropy, AdamW at a learning rate of $10^{-4}$ for the probe and $10^{-5}$ for the LoRA adapters, with LoRA rank $16$, $\alpha=32$ and dropout $0.05$, no weight decay, gradient clipping at norm $1$, $3{,}000$ steps of $4{,}096$ tokens, and three seeds. Healthy responses are subsampled to four per degenerate response, stratified by source, and a quarter of the supervised positions in each batch are positive. The loss weights positive targets by the ratio of negative to positive targets in the training set. Every probe is evaluated on the validation split every $100$ steps, and the selected strategy, \texttt{frontier\_window} centred on the onset with a window of $512$ tokens and a hard target with $K=256$, draws the supervised positions of a healthy response from a window at a random position. The strategies differ along two axes. \emph{Token selection} decides which positions are supervised. \texttt{all\_tokens} uses every position, \texttt{rollout\_balanced} samples the same number of positions from each response, \texttt{random\_window} samples one contiguous window, and \texttt{frontier\_window} places the window of a degenerate response around the onset, either centred on it or ending at it (trailing), and samples it at random in a healthy response. The \emph{target} is either hard, positive from $K$ tokens before the onset, or soft, rising towards the onset with an exponential or linear decay over $\lambda$ tokens. \Cref{tab:strategies} compares the $26$ frozen training strategies and the adapted version of the selected one. At layer $31$, the last layer, a frozen probe reads the activation after the model's final normalization, since this is the last activation the model returns. Every other layer is read at its output. All experiments run on the Alps supercomputer of the Swiss National Supercomputing Centre (CSCS), on NVIDIA GH200 GPUs.

\begin{table}[!htbp]
  \centering
  \caption{AUROC varies little between training strategies, while warning
coverage varies several-fold. Training strategies compared on
\textsc{Apertus}. Within each group, strategies are ordered by validation
$\mathrm{Warn}_{256}$ at $\beta = 1\%$, the score the selection rule of
\Cref{sec:prewarn} uses, and the selected strategy is in bold. The number
after a window strategy is its window size in tokens. Layer: median and
range over seeds of the selected layer. $\mathrm{Warn}_{256}$ (val, $1\%$)
is measured on validation at $\beta = 1\%$, the other metric columns on the
human-verified in-domain test split at $\beta = 5\%$. All values are mean
$\pm$ one standard deviation over three seeds, each read at its selected
layer.}
  \label{tab:strategies}
  \footnotesize
  \setlength{\tabcolsep}{3pt}
  \resizebox{\linewidth}{!}{%
  \begin{tabular}{lrrrrrr}
    \toprule
    Training strategy & Layer & AUROC & Post & $\mathrm{Warn}_{256}$ (val, $1\%$) & $\mathrm{Warn}_{256}$ & Early \\
    \midrule
    \multicolumn{7}{l}{\textbf{Token selection} (frozen, hard target, $K{=}0$)} \\
    \texttt{rollout\_balanced} (128) & $16$ (15--16) & $0.9998 \pm 0.0001$ & $0.925 \pm 0.001$ & $0.0482 \pm 0.0043$ & $0.158 \pm 0.004$ & $0.580 \pm 0.010$ \\
    \texttt{all\_tokens} & $12$ (12--15) & $0.9997 \pm 0.0000$ & $0.923 \pm 0.007$ & $0.0475 \pm 0.0040$ & $0.166 \pm 0.016$ & $0.552 \pm 0.023$ \\
    \texttt{random\_window} (64) & $15$ (12--15) & $0.9998 \pm 0.0000$ & $0.927 \pm 0.006$ & $0.0444 \pm 0.0017$ & $0.159 \pm 0.014$ & $0.552 \pm 0.015$ \\
    \texttt{frontier\_window} (256) & $6$ (6--16) & $0.9990 \pm 0.0002$ & $0.814 \pm 0.025$ & $0.0416 \pm 0.0051$ & $0.098 \pm 0.008$ & $0.552 \pm 0.038$ \\
    \texttt{frontier\_window} (128) & $4$ & $0.9986 \pm 0.0002$ & $0.610 \pm 0.005$ & $0.0380 \pm 0.0009$ & $0.072 \pm 0.001$ & $0.486 \pm 0.005$ \\
    \texttt{random\_window} (128) & $12$ (4--15) & $0.9993 \pm 0.0009$ & $0.867 \pm 0.092$ & $0.0370 \pm 0.0027$ & $0.127 \pm 0.037$ & $0.514 \pm 0.039$ \\
    \texttt{frontier\_window} (512) & $12$ (5--16) & $0.9987 \pm 0.0012$ & $0.823 \pm 0.115$ & $0.0363 \pm 0.0050$ & $0.111 \pm 0.034$ & $0.537 \pm 0.057$ \\
    \texttt{random\_window} (256) & $4$ (4--12) & $0.9985 \pm 0.0011$ & $0.802 \pm 0.108$ & $0.0328 \pm 0.0009$ & $0.106 \pm 0.043$ & $0.474 \pm 0.034$ \\
    \texttt{random\_window} (512) & $4$ (4--12) & $0.9981 \pm 0.0014$ & $0.777 \pm 0.122$ & $0.0321 \pm 0.0011$ & $0.097 \pm 0.039$ & $0.466 \pm 0.023$ \\
    \texttt{frontier\_window} (64) & $6$ & $0.9981 \pm 0.0004$ & $0.526 \pm 0.032$ & $0.0239 \pm 0.0006$ & $0.042 \pm 0.003$ & $0.477 \pm 0.005$ \\
    \addlinespace
    \multicolumn{7}{l}{\textbf{Target horizon} (frozen, hard target from $K$ tokens before the onset)} \\
    \textbf{frontier\_window} \textbf{(512)} \textbf{centred} $\mathbf{K{=}256}$ & $15$ & $0.9994 \pm 0.0000$ & $0.865 \pm 0.007$ & $\mathbf{0.0743 \pm 0.0045}$ & $0.186 \pm 0.007$ & $0.658 \pm 0.013$ \\
    \texttt{all\_tokens} $K{=}1024$ & $15$ (14--15) & $0.9996 \pm 0.0001$ & $0.900 \pm 0.004$ & $0.0737 \pm 0.0042$ & $0.193 \pm 0.010$ & $0.647 \pm 0.031$ \\
    \texttt{all\_tokens} $K{=}256$ & $12$ (12--15) & $0.9997 \pm 0.0000$ & $0.921 \pm 0.007$ & $0.0714 \pm 0.0009$ & $0.205 \pm 0.018$ & $0.589 \pm 0.010$ \\
    \texttt{frontier\_window} (512) trailing $K{=}512$ & $15$ & $0.9989 \pm 0.0000$ & $0.823 \pm 0.016$ & $0.0688 \pm 0.0063$ & $0.172 \pm 0.009$ & $0.670 \pm 0.005$ \\
    \texttt{frontier\_window} (1024) trailing $K{=}512$ & $15$ & $0.9987 \pm 0.0001$ & $0.858 \pm 0.007$ & $0.0592 \pm 0.0033$ & $0.162 \pm 0.008$ & $0.664 \pm 0.000$ \\
    \texttt{frontier\_window} (512) centred $K{=}128$ & $15$ (14--15) & $0.9993 \pm 0.0001$ & $0.878 \pm 0.004$ & $0.0563 \pm 0.0065$ & $0.161 \pm 0.005$ & $0.632 \pm 0.005$ \\
    \texttt{frontier\_window} (512) trailing $K{=}256$ & $15$ & $0.9985 \pm 0.0001$ & $0.826 \pm 0.018$ & $0.0548 \pm 0.0048$ & $0.143 \pm 0.012$ & $0.647 \pm 0.015$ \\
    \texttt{frontier\_window} (1024) centred $K{=}256$ & $12$ & $0.9994 \pm 0.0000$ & $0.910 \pm 0.000$ & $0.0510 \pm 0.0043$ & $0.194 \pm 0.004$ & $0.618 \pm 0.005$ \\
    \texttt{frontier\_window} (1024) trailing $K{=}256$ & $15$ (12--16) & $0.9983 \pm 0.0005$ & $0.861 \pm 0.022$ & $0.0437 \pm 0.0049$ & $0.144 \pm 0.020$ & $0.619 \pm 0.024$ \\
    \texttt{frontier\_window} (512) trailing $K{=}128$ & $12$ (8--16) & $0.9975 \pm 0.0007$ & $0.777 \pm 0.064$ & $0.0373 \pm 0.0042$ & $0.108 \pm 0.018$ & $0.545 \pm 0.020$ \\
    \texttt{frontier\_window} (1024) centred $K{=}128$ & $16$ (12--16) & $0.9992 \pm 0.0002$ & $0.895 \pm 0.011$ & $0.0372 \pm 0.0039$ & $0.143 \pm 0.015$ & $0.572 \pm 0.013$ \\
    \texttt{frontier\_window} (1024) trailing $K{=}128$ & $12$ (7--15) & $0.9977 \pm 0.0015$ & $0.825 \pm 0.071$ & $0.0289 \pm 0.0020$ & $0.109 \pm 0.030$ & $0.546 \pm 0.070$ \\
    \addlinespace
    \multicolumn{7}{l}{\textbf{Soft targets} (frozen, \texttt{frontier\_window} (512) centred)} \\
    exponential, decay $256$ & $15$ (14--17) & $0.9992 \pm 0.0002$ & $0.877 \pm 0.009$ & $0.0577 \pm 0.0074$ & $0.165 \pm 0.003$ & $0.644 \pm 0.013$ \\
    linear, decay $256$ & $15$ & $0.9994 \pm 0.0001$ & $0.876 \pm 0.002$ & $0.0504 \pm 0.0083$ & $0.152 \pm 0.005$ & $0.621 \pm 0.015$ \\
    exponential, decay $128$ & $12$ (5--15) & $0.9989 \pm 0.0011$ & $0.823 \pm 0.110$ & $0.0460 \pm 0.0061$ & $0.134 \pm 0.044$ & $0.569 \pm 0.075$ \\
    linear, decay $128$ & $5$ (5--15) & $0.9982 \pm 0.0012$ & $0.735 \pm 0.107$ & $0.0384 \pm 0.0008$ & $0.092 \pm 0.026$ & $0.497 \pm 0.070$ \\
    \addlinespace
    \multicolumn{7}{l}{\textbf{Adapted} (LoRA rank 16 on layers $0$ to $\ell$)} \\
    \texttt{frontier\_window} (512) centred $K{=}256$ & $30$ & $0.9982 \pm 0.0012$ & $0.973 \pm 0.012$ & $0.3035 \pm 0.0163$ & $0.577 \pm 0.013$ & $0.716 \pm 0.015$ \\
    \bottomrule
  \end{tabular}}
\end{table}

\FloatBarrier
\subsection{Response-level against onset-aligned metrics}
\label{app:ranking}

\paragraph{Scorers.}
The scorers of \Cref{fig:metrics} are, at layer $15$ of \textsc{Apertus}, the
frozen \texttt{random\_window} probe with a window of $64$ tokens and $K{=}0$, which has the
highest validation AUROC of the $26$ frozen training strategies, the frozen
\texttt{frontier\_window} probe with a trailing window of $1{,}024$ tokens and
$K{=}128$, which has the lowest among the strategies whose three seeds all
pass the post-onset floor at this layer, and the adapted probe. Panel (d) reads the
frozen probe with the training strategy of \Cref{sec:prewarn}, which the adapted
probe also uses.

\paragraph{Same AUROC, different timing.}
The four scorers of \Cref{fig:metrics} reach almost the same AUROC (0.9980 to 0.9998) but differ several-fold in warning coverage.
A response-level metric scores each response only by its highest token score, $\max_t s_t$, so it cannot see when a scorer fires.
\Cref{tab:positional-control} shows how far apart two scorers with the same AUROC can be.
From each probe we build two scorers that keep the highest score of every response, and so also its AUROC and calibrated threshold at persistence $m=1$, but move its other flags to opposite extremes.
\emph{Shuffled} permutes the scores within each response, which spreads the flags over the whole response, including its first tokens.
\emph{Argmax only} sets every score except the highest to the minimum, which leaves at most one flag per response, usually deep inside the loop.
\Cref{tab:positional-control} is computed at $\beta=1\%$.

With the same AUROC, warning coverage ranges from 0.000 for argmax only to 0.52-0.67 for shuffled, and the early-warning rate from 0.01-0.02 to 0.95-0.96.
Neither is a monitor one could deploy, since shuffling moves scores from after the onset to the start of the response, but both have the AUROC of a real probe.
A response-level metric therefore cannot tell a scorer that never warns from one that warns before almost every loop.
With persistence $m=4$ and the threshold recalibrated, shuffled keeps its AUROC and its early alarms, while argmax only never fires (AUROC 0.5).

\paragraph{Selection by response-level metrics.}
Because the response-level metrics do not see where the flags fall, selecting
a probe by them can pick one that warns poorly. Ranked by validation AUROC, the
adapted probe is $14$th of the $27$ probes at layer $15$, $0.0003$ below the
top. Selecting by validation AUROC picks the \texttt{random\_window} probe,
whose test warning coverage is $3.4\times$ lower than the adapted probe's.
Among the frozen probes alone, the probe selected by AUROC covers $1.2\times$
less of the warning window than the one selected by warning coverage.
\Cref{tab:rank-transfer} compares the ranking of the $27$ probes, each at its
selected layer, on validation and on test (\Cref{sec:rq-metrics} discusses
it).

\begin{table}[!htbp]
  \centering
  \caption{Scorers with the same AUROC range from one that almost never warns
to one that warns before almost every loop. From each probe of
\Cref{fig:metrics}, two transforms keep the highest score of every response,
and with it the AUROC and the calibrated threshold at $m = 1$, but move its
other flags. Shuffled: the token scores of each response permuted at
random, which spreads the flags over the whole response. Argmax only: every
score except the highest set to the minimum, which leaves at most one flag
per response. As scored: the original scores. $m$: number of consecutive
flagged tokens needed for an alarm (\Cref{eq:alarm}); at $m = 4$ the
threshold is recalibrated. AUROC is given once per scorer for $m = 1$.
--: no alarm. Lead in tokens, negative for warnings. \textsc{Apertus},
layer~$15$, human-verified in-domain test split, $\beta = 1\%$, mean over
three seeds}
  \label{tab:positional-control}
  \footnotesize
  \setlength{\tabcolsep}{3pt}
  \resizebox{\linewidth}{!}{%
  \begin{tabular}{ll rrrrr rrrrrr}
    \toprule
    & & \multicolumn{5}{c}{$m=1$} & \multicolumn{6}{c}{$m=4$} \\
    \cmidrule(lr){3-7}\cmidrule(lr){8-13}
    Scorer & Scores & Post & $\mathrm{Warn}_{256}$ & Early & Lead & Pre & AUROC & Post & $\mathrm{Warn}_{256}$ & Early & Lead & Pre \\
    \midrule
    \multirow{3}{*}{\shortstack[l]{frozen, highest AUROC\\AUROC $0.9998$}} & as scored & 0.822 & 0.028 & 0.24 & $+124$ & 0.018 & 0.9998 & 0.883 & 0.080 & 0.24 & $+124$ & 0.044 \\
     & shuffled & 0.678 & 0.627 & 0.96 & $-816$ & 0.507 & 0.9997 & 0.788 & 0.753 & 0.96 & $-809$ & 0.679 \\
     & argmax only & 0.0003 & 0.000 & 0.01 & $+2694$ & 0.00001 & 0.5000 & -- & -- & -- & -- & -- \\
    \midrule
    \multirow{3}{*}{\shortstack[l]{frozen, lowest AUROC\\AUROC $0.9980$}} & as scored & 0.698 & 0.018 & 0.24 & $+88$ & 0.001 & 0.9973 & 0.792 & 0.052 & 0.20 & $+110$ & 0.009 \\
     & shuffled & 0.585 & 0.518 & 0.95 & $-795$ & 0.377 & 0.9971 & 0.752 & 0.703 & 0.95 & $-791$ & 0.597 \\
     & argmax only & 0.0003 & 0.000 & 0.01 & $+2037$ & 0.00001 & 0.5000 & -- & -- & -- & -- & -- \\
    \midrule
    \multirow{3}{*}{\shortstack[l]{adapted\\AUROC $0.9991$}} & as scored & 0.846 & 0.220 & 0.43 & $+41$ & 0.111 & 0.9990 & 0.899 & 0.296 & 0.43 & $+33$ & 0.166 \\
     & shuffled & 0.714 & 0.671 & 0.96 & $-819$ & 0.600 & 0.9992 & 0.830 & 0.802 & 0.96 & $-810$ & 0.760 \\
     & argmax only & 0.0003 & 0.00004 & 0.02 & $+2294$ & 0.00001 & 0.5000 & -- & -- & -- & -- & -- \\
    \bottomrule
  \end{tabular}%
  }
\end{table}

\begin{table}[!htbp]
  \centering
  \caption{Rank transfer from validation to the human-verified in-domain test split, Kendall's
  $\tau$ over the $27$ \textsc{Apertus} training strategies of \Cref{tab:strategies}, each read at
  the layers selected there, at $\beta=5\%$. Left: whether a metric
  reproduces its own ordering. Right: whether its validation ordering predicts
  the test ordering by warning coverage.}
  \label{tab:rank-transfer}
  \small
  \begin{tabular}{lcc}
    \toprule
    Ranking on validation & Reproduces itself on test & Predicts test $\mathrm{Warn}_{256}$ \\
    \midrule
    AUROC & $+0.79$ & $+0.33$ \\
    Best accuracy & $+0.53$ & $+0.36$ \\
    \midrule
    $\mathrm{Warn}_{256}$ & $+0.94$ & $\mathbf{+0.94}$ \\
    Post & $+0.93$ & $+0.61$ \\
    \bottomrule
  \end{tabular}
\end{table}

\FloatBarrier
\subsection{Baselines}
\label{app:baselines}

\paragraph{Definitions.}
Every baseline assigns a score $s_t\in[0,1]$ to each position $t$ of a response
$r=(r_1,\ldots,r_T)$ using only $r_1,\ldots,r_t$ and the generator's output up to
$t$. Only the order of the scores matters, since every threshold is calibrated
on validation. Several baselines average over the trailing window
\[
\mathcal{S}_t=\{\max(1,t-S+1),\ldots,t\},
\]
which holds fewer than $S$ positions at the start of a response, so that a
short prefix is averaged over the tokens it has.

\emph{Token-cap rule.} The score is $1$ on the last token of a capped response
and $0$ elsewhere,
\[
s_t=\mathbf{1}\left[t=T=T_{\max}\right].
\]

\emph{Windowed seq-rep-$n$.} Let $G_t$ be the list of $n$-grams that lie
entirely inside the window $\mathcal{S}_t$. The score is the share of these
$n$-grams that repeat an earlier one in the same window \citep{welleck2020neural},
\[
s_t=1-\frac{\left|\operatorname{set}(G_t)\right|}{\left|G_t\right|},
\qquad t\geq n,
\]
and $s_t=0$ for $t<n$, where no complete $n$-gram exists yet.

\emph{rep$/\ell$.} Here $\ell$ is a number of tokens, not a layer. A token is
counted as repeated if it already occurred among the $\ell$ tokens before it,
\[
z_t=\mathbf{1}\left[r_t\in\{r_j:\max(1,t-\ell)\leq j<t\}\right],
\qquad
s_t=\frac{1}{|\mathcal{S}_t|}\sum_{j\in\mathcal{S}_t}z_j .
\]

\emph{Repeated-$n$-gram rule.} Let $g_t=(r_{t-n+1},\ldots,r_t)$ be the $n$-gram
ending at $t$ and $c_t=\left|\{j: n\leq j\leq t,\ g_j=g_t\}\right|$ the number of
times it has occurred so far, including at $t$. The score is
\[
s_t=1-\frac{1}{c_t},\qquad t\geq n,
\]
and $s_t=0$ for $t<n$. A threshold $\tau$ on $s_t$ is a threshold
$1/(1-\tau)$ on the count, so calibrating $\tau$ chooses how many repeats of an
$n$-gram trigger an alarm. \citet{pipis2025wait} use a count of this kind to
label a response as looping, when some $30$-gram occurs at least $20$ times.

\emph{Windowed entropy.} Let $p_t$ be the generator's next-token distribution
at position $t$, computed from the raw logits at temperature $1$, before
top-$p$ truncation, and $H_t=-\sum_{v}p_t(v)\log p_t(v)$ its entropy. The score
inverts the average entropy over the window,
\[
s_t=\frac{1}{1+\frac{1}{|\mathcal{S}_t|}\sum_{j\in\mathcal{S}_t}H_j},
\]
so a stretch of confident predictions scores close to $1$.

\emph{Windowed negative log-probability.} The same construction on the surprise
of the emitted token, $u_t=-\log p_t(r_t)$,
\[
s_t=\frac{1}{1+\frac{1}{|\mathcal{S}_t|}\sum_{j\in\mathcal{S}_t}u_j}.
\]

\emph{Random alarms.} Each token is flagged independently with probability
$q$, chosen so that the expected share of healthy validation responses that
raise an alarm equals the budget,
\[
\frac{1}{|\mathcal{H}|}\sum_{r\in\mathcal{H}}\left(1-(1-q)^{T_r}\right)=\beta .
\]
Their metrics follow in closed form from the lengths and onsets of the responses.

\paragraph{Selection.}
The window $S$, the look-back $\ell$ and the $n$-gram size $n$ are chosen per
model on the validation split with the rule that selects the probes' layer and
checkpoint. Among the settings whose post-onset coverage at $\beta=1\%$ is at
least $0.3$, we take the one with the highest $\mathrm{Warn}_{256}$ at
$\beta=1\%$. \Cref{tab:sweep-confidence,tab:sweep-ngram,tab:sweep-seqrep,tab:sweep-repl}
list every setting we evaluated. For windowed entropy and windowed negative
log-probability on \textsc{Llama}, no setting reaches the floor, and we take the
one with the highest $\mathrm{Warn}_{256}$. Windowed seq-rep-$n$ and rep$/\ell$
select windows of $1{,}024$ to $4{,}096$ tokens, the confidence-based baselines
windows of $256$ and $512$ tokens.

\begin{table}[!htbp]
  \centering
  \caption{Window $S$ of windowed entropy and windowed negative log-probability. On \textsc{Llama} no window reaches the floor, and the window with the highest $\mathrm{Warn}_{256}$ is selected. Validation split, $\beta=1\%$. Bold: the setting selected for that model (\Cref{app:baselines}). Grey: below the post-onset floor of $0.3$. A coverage near $0$ marks a setting on which healthy responses already reach the top of the score range, so the calibrated threshold sits at that top and almost no token is flagged.}
  \label{tab:sweep-confidence}
  \scriptsize
  \setlength{\tabcolsep}{4pt}
  \begin{tabular}{lrrrrrrr}
    \toprule
     &  & \multicolumn{2}{c}{\textsc{Apertus}} & \multicolumn{2}{c}{\textsc{Llama}} & \multicolumn{2}{c}{\textsc{Mistral}} \\
    \cmidrule(lr){3-4}\cmidrule(lr){5-6}\cmidrule(lr){7-8}
    Baseline & $S$ & Post & $\mathrm{Warn}_{256}$ & Post & $\mathrm{Warn}_{256}$ & Post & $\mathrm{Warn}_{256}$ \\
    \midrule
    entropy & 8 & \textcolor{gray}{0.000} & \textcolor{gray}{0.000} & \textcolor{gray}{0.000} & \textcolor{gray}{0.000} & \textcolor{gray}{0.000} & \textcolor{gray}{0.000} \\
     & 16 & \textcolor{gray}{0.002} & \textcolor{gray}{0.000} & \textcolor{gray}{0.000} & \textcolor{gray}{0.001} & \textcolor{gray}{0.000} & \textcolor{gray}{0.000} \\
     & 32 & \textcolor{gray}{0.008} & \textcolor{gray}{0.000} & \textcolor{gray}{0.001} & \textcolor{gray}{0.002} & \textcolor{gray}{0.000} & \textcolor{gray}{0.000} \\
     & 64 & \textcolor{gray}{0.018} & \textcolor{gray}{0.003} & \textcolor{gray}{0.001} & \textcolor{gray}{0.002} & \textcolor{gray}{0.000} & \textcolor{gray}{0.000} \\
     & 128 & \textcolor{gray}{0.228} & \textcolor{gray}{0.005} & \textcolor{gray}{0.003} & \textcolor{gray}{0.003} & \textcolor{gray}{0.048} & \textcolor{gray}{0.003} \\
     & 256 & 0.671 & 0.010 & \textcolor{gray}{\textbf{0.018}} & \textcolor{gray}{\textbf{0.010}} & 0.349 & 0.008 \\
     & 512 & \textbf{0.667} & \textbf{0.017} & \textcolor{gray}{0.024} & \textcolor{gray}{0.007} & \textbf{0.616} & \textbf{0.014} \\
     & 1024 & 0.612 & 0.001 & \textcolor{gray}{0.035} & \textcolor{gray}{0.003} & 0.549 & 0.007 \\
     & 2048 & 0.368 & 0.000 & \textcolor{gray}{0.024} & \textcolor{gray}{0.000} & 0.369 & 0.003 \\
    \midrule
    $-\log p$ & 8 & \textcolor{gray}{0.000} & \textcolor{gray}{0.000} & \textcolor{gray}{0.000} & \textcolor{gray}{0.000} & \textcolor{gray}{0.000} & \textcolor{gray}{0.000} \\
     & 16 & \textcolor{gray}{0.000} & \textcolor{gray}{0.000} & \textcolor{gray}{0.000} & \textcolor{gray}{0.001} & \textcolor{gray}{0.000} & \textcolor{gray}{0.000} \\
     & 32 & \textcolor{gray}{0.007} & \textcolor{gray}{0.000} & \textcolor{gray}{0.001} & \textcolor{gray}{0.002} & \textcolor{gray}{0.000} & \textcolor{gray}{0.000} \\
     & 64 & \textcolor{gray}{0.020} & \textcolor{gray}{0.003} & \textcolor{gray}{0.001} & \textcolor{gray}{0.002} & \textcolor{gray}{0.000} & \textcolor{gray}{0.000} \\
     & 128 & \textcolor{gray}{0.229} & \textcolor{gray}{0.005} & \textcolor{gray}{0.003} & \textcolor{gray}{0.005} & \textcolor{gray}{0.067} & \textcolor{gray}{0.005} \\
     & 256 & \textbf{0.665} & \textbf{0.010} & \textcolor{gray}{\textbf{0.011}} & \textcolor{gray}{\textbf{0.009}} & \textbf{0.408} & \textbf{0.010} \\
     & 512 & 0.665 & 0.005 & \textcolor{gray}{0.027} & \textcolor{gray}{0.006} & 0.616 & 0.007 \\
     & 1024 & 0.588 & 0.000 & \textcolor{gray}{0.033} & \textcolor{gray}{0.000} & 0.530 & 0.007 \\
     & 2048 & 0.336 & 0.000 & \textcolor{gray}{0.010} & \textcolor{gray}{0.000} & 0.318 & 0.003 \\
    \bottomrule
  \end{tabular}
\end{table}

\begin{table}[!htbp]
  \centering
  \caption{$N$-gram size $n$ of the repeated-$n$-gram rule. Validation split, $\beta=1\%$. Bold: the setting selected for that model (\Cref{app:baselines}). Grey: below the post-onset floor of $0.3$. A coverage near $0$ marks a setting on which healthy responses already reach the top of the score range, so the calibrated threshold sits at that top and almost no token is flagged.}
  \label{tab:sweep-ngram}
  \scriptsize
  \setlength{\tabcolsep}{4pt}
  \begin{tabular}{rrrrrrr}
    \toprule
     & \multicolumn{2}{c}{\textsc{Apertus}} & \multicolumn{2}{c}{\textsc{Llama}} & \multicolumn{2}{c}{\textsc{Mistral}} \\
    \cmidrule(lr){2-3}\cmidrule(lr){4-5}\cmidrule(lr){6-7}
    $n$ & Post & $\mathrm{Warn}_{256}$ & Post & $\mathrm{Warn}_{256}$ & Post & $\mathrm{Warn}_{256}$ \\
    \midrule
    4 & 0.426 & 0.006 & \textcolor{gray}{0.178} & \textcolor{gray}{0.006} & 0.446 & 0.009 \\
    6 & 0.486 & 0.008 & \textcolor{gray}{0.226} & \textcolor{gray}{0.007} & 0.512 & 0.008 \\
    8 & 0.542 & 0.007 & \textcolor{gray}{0.268} & \textcolor{gray}{0.008} & 0.527 & 0.006 \\
    12 & 0.632 & 0.011 & \textcolor{gray}{0.281} & \textcolor{gray}{0.006} & 0.605 & 0.006 \\
    16 & 0.643 & 0.008 & 0.334 & 0.006 & 0.634 & 0.007 \\
    24 & \textbf{0.733} & \textbf{0.022} & 0.363 & 0.005 & 0.665 & 0.007 \\
    30 & 0.720 & 0.015 & 0.416 & 0.007 & 0.667 & 0.007 \\
    48 & 0.738 & 0.010 & 0.477 & 0.008 & 0.716 & 0.009 \\
    64 & 0.775 & 0.014 & 0.499 & 0.007 & \textbf{0.721} & \textbf{0.010} \\
    96 & 0.752 & 0.007 & \textbf{0.523} & \textbf{0.011} & 0.722 & 0.008 \\
    128 & 0.734 & 0.003 & 0.502 & 0.006 & 0.735 & 0.008 \\
    \bottomrule
  \end{tabular}
\end{table}

\begin{table}[!htbp]
  \centering
  \caption{$N$-gram order $n$ and window $S$ of windowed seq-rep-$n$. Validation split, $\beta=1\%$. Bold: the setting selected for that model (\Cref{app:baselines}). Grey: below the post-onset floor of $0.3$. A coverage near $0$ marks a setting on which healthy responses already reach the top of the score range, so the calibrated threshold sits at that top and almost no token is flagged.}
  \label{tab:sweep-seqrep}
  \scriptsize
  \setlength{\tabcolsep}{4pt}
  \begin{tabular}{rrrrrrrr}
    \toprule
     &  & \multicolumn{2}{c}{\textsc{Apertus}} & \multicolumn{2}{c}{\textsc{Llama}} & \multicolumn{2}{c}{\textsc{Mistral}} \\
    \cmidrule(lr){3-4}\cmidrule(lr){5-6}\cmidrule(lr){7-8}
    $n$ & $S$ & Post & $\mathrm{Warn}_{256}$ & Post & $\mathrm{Warn}_{256}$ & Post & $\mathrm{Warn}_{256}$ \\
    \midrule
    2 & 32 & \textcolor{gray}{0.112} & \textcolor{gray}{0.000} & \textcolor{gray}{0.028} & \textcolor{gray}{0.003} & \textcolor{gray}{0.112} & \textcolor{gray}{0.003} \\
    2 & 64 & \textcolor{gray}{0.174} & \textcolor{gray}{0.000} & \textcolor{gray}{0.061} & \textcolor{gray}{0.005} & \textcolor{gray}{0.162} & \textcolor{gray}{0.003} \\
    2 & 128 & \textcolor{gray}{0.275} & \textcolor{gray}{0.022} & \textcolor{gray}{0.144} & \textcolor{gray}{0.015} & 0.324 & 0.009 \\
    2 & 256 & 0.433 & 0.031 & \textcolor{gray}{0.241} & \textcolor{gray}{0.020} & 0.556 & 0.018 \\
    2 & 512 & 0.619 & 0.044 & 0.429 & 0.048 & 0.671 & 0.021 \\
    2 & 1024 & 0.782 & 0.099 & 0.572 & 0.065 & 0.758 & 0.032 \\
    2 & 2048 & 0.791 & 0.092 & 0.574 & 0.069 & 0.777 & 0.037 \\
    2 & 4096 & 0.780 & 0.092 & 0.516 & 0.067 & 0.774 & 0.037 \\
    3 & 32 & \textcolor{gray}{0.122} & \textcolor{gray}{0.000} & \textcolor{gray}{0.042} & \textcolor{gray}{0.004} & \textcolor{gray}{0.124} & \textcolor{gray}{0.003} \\
    3 & 64 & \textcolor{gray}{0.195} & \textcolor{gray}{0.000} & \textcolor{gray}{0.074} & \textcolor{gray}{0.007} & \textcolor{gray}{0.208} & \textcolor{gray}{0.007} \\
    3 & 128 & 0.307 & 0.020 & \textcolor{gray}{0.154} & \textcolor{gray}{0.016} & 0.364 & 0.010 \\
    3 & 256 & 0.507 & 0.043 & \textcolor{gray}{0.272} & \textcolor{gray}{0.021} & 0.565 & 0.014 \\
    3 & 512 & 0.675 & 0.049 & 0.459 & 0.046 & 0.697 & 0.019 \\
    3 & 1024 & \textbf{0.799} & \textbf{0.100} & 0.618 & 0.072 & 0.747 & 0.026 \\
    3 & 2048 & 0.810 & 0.092 & 0.634 & 0.067 & 0.789 & 0.034 \\
    3 & 4096 & 0.802 & 0.092 & 0.592 & 0.069 & \textbf{0.792} & \textbf{0.038} \\
    4 & 32 & \textcolor{gray}{0.123} & \textcolor{gray}{0.000} & \textcolor{gray}{0.043} & \textcolor{gray}{0.004} & \textcolor{gray}{0.132} & \textcolor{gray}{0.003} \\
    4 & 64 & \textcolor{gray}{0.196} & \textcolor{gray}{0.000} & \textcolor{gray}{0.084} & \textcolor{gray}{0.008} & \textcolor{gray}{0.245} & \textcolor{gray}{0.007} \\
    4 & 128 & 0.320 & 0.015 & \textcolor{gray}{0.164} & \textcolor{gray}{0.016} & 0.389 & 0.009 \\
    4 & 256 & 0.509 & 0.044 & \textcolor{gray}{0.287} & \textcolor{gray}{0.021} & 0.567 & 0.012 \\
    4 & 512 & 0.678 & 0.042 & 0.480 & 0.048 & 0.712 & 0.017 \\
    4 & 1024 & 0.813 & 0.091 & 0.639 & 0.072 & 0.756 & 0.027 \\
    4 & 2048 & 0.805 & 0.080 & 0.642 & 0.062 & 0.793 & 0.036 \\
    4 & 4096 & 0.798 & 0.080 & \textbf{0.628} & \textbf{0.072} & 0.785 & 0.033 \\
    \bottomrule
  \end{tabular}
\end{table}

\begin{table}[!htbp]
  \centering
  \caption{Look-back $\ell$ and window $S$ of rep$/\ell$. Validation split, $\beta=1\%$. Bold: the setting selected for that model (\Cref{app:baselines}). Grey: below the post-onset floor of $0.3$. A coverage near $0$ marks a setting on which healthy responses already reach the top of the score range, so the calibrated threshold sits at that top and almost no token is flagged.}
  \label{tab:sweep-repl}
  \scriptsize
  \setlength{\tabcolsep}{4pt}
  \begin{tabular}{rrrrrrrr}
    \toprule
     &  & \multicolumn{2}{c}{\textsc{Apertus}} & \multicolumn{2}{c}{\textsc{Llama}} & \multicolumn{2}{c}{\textsc{Mistral}} \\
    \cmidrule(lr){3-4}\cmidrule(lr){5-6}\cmidrule(lr){7-8}
    $\ell$ & $S$ & Post & $\mathrm{Warn}_{256}$ & Post & $\mathrm{Warn}_{256}$ & Post & $\mathrm{Warn}_{256}$ \\
    \midrule
    32 & 32 & \textcolor{gray}{0.000} & \textcolor{gray}{0.000} & \textcolor{gray}{0.000} & \textcolor{gray}{0.000} & \textcolor{gray}{0.000} & \textcolor{gray}{0.000} \\
    32 & 64 & \textcolor{gray}{0.185} & \textcolor{gray}{0.000} & \textcolor{gray}{0.110} & \textcolor{gray}{0.005} & \textcolor{gray}{0.000} & \textcolor{gray}{0.000} \\
    32 & 128 & \textcolor{gray}{0.202} & \textcolor{gray}{0.009} & \textcolor{gray}{0.119} & \textcolor{gray}{0.004} & 0.323 & 0.006 \\
    32 & 256 & \textcolor{gray}{0.212} & \textcolor{gray}{0.019} & \textcolor{gray}{0.148} & \textcolor{gray}{0.022} & 0.365 & 0.010 \\
    32 & 512 & \textcolor{gray}{0.226} & \textcolor{gray}{0.025} & \textcolor{gray}{0.178} & \textcolor{gray}{0.036} & 0.437 & 0.021 \\
    32 & 1024 & \textcolor{gray}{0.235} & \textcolor{gray}{0.026} & \textcolor{gray}{0.187} & \textcolor{gray}{0.035} & 0.440 & 0.018 \\
    32 & 2048 & \textcolor{gray}{0.221} & \textcolor{gray}{0.027} & \textcolor{gray}{0.165} & \textcolor{gray}{0.016} & 0.421 & 0.015 \\
    32 & 4096 & \textcolor{gray}{0.215} & \textcolor{gray}{0.027} & \textcolor{gray}{0.136} & \textcolor{gray}{0.016} & 0.406 & 0.011 \\
    64 & 32 & \textcolor{gray}{0.000} & \textcolor{gray}{0.000} & \textcolor{gray}{0.000} & \textcolor{gray}{0.000} & \textcolor{gray}{0.000} & \textcolor{gray}{0.000} \\
    64 & 64 & \textcolor{gray}{0.000} & \textcolor{gray}{0.000} & \textcolor{gray}{0.000} & \textcolor{gray}{0.000} & \textcolor{gray}{0.000} & \textcolor{gray}{0.000} \\
    64 & 128 & 0.325 & 0.019 & \textcolor{gray}{0.165} & \textcolor{gray}{0.006} & 0.493 & 0.013 \\
    64 & 256 & 0.344 & 0.021 & \textcolor{gray}{0.231} & \textcolor{gray}{0.027} & 0.517 & 0.017 \\
    64 & 512 & 0.390 & 0.038 & \textcolor{gray}{0.268} & \textcolor{gray}{0.043} & 0.565 & 0.024 \\
    64 & 1024 & 0.403 & 0.031 & \textcolor{gray}{0.286} & \textcolor{gray}{0.052} & 0.574 & 0.024 \\
    64 & 2048 & 0.381 & 0.032 & \textcolor{gray}{0.254} & \textcolor{gray}{0.029} & 0.543 & 0.016 \\
    64 & 4096 & 0.359 & 0.032 & \textcolor{gray}{0.204} & \textcolor{gray}{0.026} & 0.529 & 0.013 \\
    128 & 32 & \textcolor{gray}{0.000} & \textcolor{gray}{0.000} & \textcolor{gray}{0.000} & \textcolor{gray}{0.000} & \textcolor{gray}{0.000} & \textcolor{gray}{0.000} \\
    128 & 64 & \textcolor{gray}{0.000} & \textcolor{gray}{0.000} & \textcolor{gray}{0.000} & \textcolor{gray}{0.000} & \textcolor{gray}{0.000} & \textcolor{gray}{0.000} \\
    128 & 128 & \textcolor{gray}{0.000} & \textcolor{gray}{0.000} & \textcolor{gray}{0.000} & \textcolor{gray}{0.000} & \textcolor{gray}{0.000} & \textcolor{gray}{0.000} \\
    128 & 256 & 0.429 & 0.014 & 0.307 & 0.013 & 0.633 & 0.021 \\
    128 & 512 & 0.497 & 0.023 & 0.377 & 0.042 & 0.647 & 0.023 \\
    128 & 1024 & 0.535 & 0.038 & 0.411 & 0.047 & 0.679 & 0.021 \\
    128 & 2048 & 0.512 & 0.033 & 0.366 & 0.039 & 0.644 & 0.017 \\
    128 & 4096 & 0.489 & 0.034 & \textcolor{gray}{0.262} & \textcolor{gray}{0.026} & 0.653 & 0.022 \\
    256 & 32 & \textcolor{gray}{0.000} & \textcolor{gray}{0.000} & \textcolor{gray}{0.000} & \textcolor{gray}{0.000} & \textcolor{gray}{0.000} & \textcolor{gray}{0.000} \\
    256 & 64 & \textcolor{gray}{0.000} & \textcolor{gray}{0.000} & \textcolor{gray}{0.000} & \textcolor{gray}{0.000} & \textcolor{gray}{0.000} & \textcolor{gray}{0.000} \\
    256 & 128 & \textcolor{gray}{0.000} & \textcolor{gray}{0.000} & \textcolor{gray}{0.000} & \textcolor{gray}{0.000} & \textcolor{gray}{0.000} & \textcolor{gray}{0.000} \\
    256 & 256 & 0.615 & 0.014 & 0.398 & 0.007 & \textcolor{gray}{0.000} & \textcolor{gray}{0.000} \\
    256 & 512 & 0.675 & 0.029 & 0.464 & 0.031 & 0.695 & 0.020 \\
    256 & 1024 & 0.721 & 0.048 & 0.511 & 0.047 & 0.749 & 0.029 \\
    256 & 2048 & 0.680 & 0.032 & 0.481 & 0.064 & 0.719 & 0.034 \\
    256 & 4096 & 0.644 & 0.035 & 0.340 & 0.039 & 0.696 & 0.026 \\
    512 & 256 & \textcolor{gray}{0.000} & \textcolor{gray}{0.000} & \textcolor{gray}{0.000} & \textcolor{gray}{0.000} & \textcolor{gray}{0.000} & \textcolor{gray}{0.000} \\
    512 & 512 & 0.807 & 0.051 & 0.485 & 0.021 & 0.765 & 0.029 \\
    512 & 1024 & 0.795 & 0.086 & 0.565 & 0.043 & 0.773 & 0.041 \\
    512 & 2048 & 0.774 & 0.083 & 0.547 & 0.063 & 0.752 & 0.040 \\
    512 & 4096 & 0.745 & 0.069 & 0.428 & 0.055 & 0.749 & 0.034 \\
    1024 & 256 & \textcolor{gray}{0.000} & \textcolor{gray}{0.000} & \textcolor{gray}{0.000} & \textcolor{gray}{0.000} & \textcolor{gray}{0.000} & \textcolor{gray}{0.000} \\
    1024 & 512 & 0.825 & 0.034 & 0.524 & 0.030 & \textcolor{gray}{0.000} & \textcolor{gray}{0.000} \\
    1024 & 1024 & \textbf{0.801} & \textbf{0.090} & 0.582 & 0.066 & 0.766 & 0.032 \\
    1024 & 2048 & 0.779 & 0.082 & \textbf{0.593} & \textbf{0.083} & \textbf{0.759} & \textbf{0.042} \\
    1024 & 4096 & 0.767 & 0.075 & 0.444 & 0.062 & 0.746 & 0.039 \\
    \bottomrule
  \end{tabular}
\end{table}

\FloatBarrier
\subsection{Judge-labelled and human-verified test splits}
\label{app:labels}

The main text reports the human-verified in-domain test split. \Cref{tab:probes-judge}
gives the same scorers on the judge-labelled in-domain test split, before
review. \Cref{tab:heldout,tab:heldout-human} report them on the two held-out
sources, excluded from training and validation, with judge labels and with
human-verified labels. All thresholds are calibrated on the in-domain
validation split. \Cref{tab:per-source} splits warning coverage and the
early-warning rate by source, for both test splits.

\begin{table}[!htbp]
  \centering
  \caption{As \Cref{tab:baselines}, with judge labels.}
  \label{tab:probes-judge}
  \footnotesize
  \setlength{\tabcolsep}{3pt}
  \resizebox{\linewidth}{!}{%
  \begin{tabular}{@{}lrrrrrrrr@{}}
    \toprule
    & \shortstack{Response\\classification} & \multicolumn{2}{c}{Failure detection} & \multicolumn{3}{c}{Early warning} & & \\
    \cmidrule(lr){2-2}\cmidrule(lr){3-4}\cmidrule(lr){5-7}
    Scorer & AUROC $\uparrow$ & Detection $\uparrow$ & Post $\uparrow$ & $\mathrm{Warn}_{256}$ $\uparrow$ & Early $\uparrow$ & Lead $\downarrow$ & FA (\%) & Pre \\
    \midrule
    \multicolumn{9}{@{}l}{\textsc{Apertus} ($n=117$)} \\
    \quad random alarms & $0.8902$ & $0.32$ & $1\times10^{-4}$ & $1\times10^{-4}$ & $0.248$$^\dagger$ & $+971$$^\dagger$ & $5.0$ & $1\times10^{-4}$ \\
    \quad token-cap rule & $0.9991$ & $1.00$ & $3\times10^{-4}$ & $0.000$ & $0.000$ & $+3261$ & $0.2$ & $0.000$ \\
    \addlinespace[2pt]
    \quad repeated-$n$-gram rule & $0.9916$ & $0.98$ & $0.774$ & $0.008$ & $0.148$ & $+147$ & $3.2$ & $0.002$ \\
    \quad windowed seq-rep-$n$ & $0.9815$ & $0.93$ & $0.880$ & $0.231$ & $0.312$ & $+141$ & $5.2$ & $0.120$ \\
    \quad rep$/\ell$ & $0.9985$ & $1.00$ & $0.916$ & $0.265$ & $0.308$ & $+201$ & $4.7$ & $0.140$ \\
    \addlinespace[2pt]
    \quad windowed entropy & $0.9971$ & $1.00$ & $0.795$ & $0.032$ & $0.120$ & $+526$ & $6.1$ & $0.033$ \\
    \quad \begin{tabular}[t]{@{}l@{}}windowed negative\\ log-probability\end{tabular} & $0.9964$ & $1.00$ & $0.813$ & $0.040$ & $0.145$ & $+383$ & $5.2$ & $0.033$ \\
    \addlinespace[2pt]
    \quad frozen probe, L15 & $\mathbf{0.9993\sd{0.0000}}$ & $1.00\sd{0.00}$ & $0.858\sd{0.007}$ & $0.237\sd{0.006}$ & $0.661\sd{0.005}$ & $-190\sd{8}$ & $5.6\sd{0.3}$ & $0.060\sd{0.003}$ \\
    \quad adapted probe, L15 & $0.9991\sd{0.0001}$ & $1.00\sd{0.00}$ & \underline{$0.966\sd{0.002}$} & $0.570\sd{0.017}$ & $\mathbf{0.732\sd{0.013}}$ & \underline{$-258\sd{14}$} & $4.8\sd{0.6}$ & $0.333\sd{0.011}$ \\
    \quad adapted probe, L30 & \underline{$0.9982\sd{0.0013}$} & $1.00\sd{0.00}$ & $\mathbf{0.973\sd{0.013}}$ & $\mathbf{0.607\sd{0.023}}$ & \underline{$0.724\sd{0.026}$} & $\mathbf{-269\sd{23}}$ & $4.6\sd{0.1}$ & $0.389\sd{0.022}$ \\
    \midrule
    \multicolumn{9}{@{}l}{\textsc{Llama} ($n=291$)} \\
    \quad random alarms & $0.8818$ & $0.29$ & $9\times10^{-5}$ & $9\times10^{-5}$ & $0.341$$^\dagger$ & $+627$$^\dagger$ & $4.8$ & $9\times10^{-5}$ \\
    \quad token-cap rule & $\mathbf{0.9999}$ & $1.00$ & $4\times10^{-4}$ & $0.000$ & $0.000$ & $+2964$ & $0.0$ & $0.000$ \\
    \addlinespace[2pt]
    \quad repeated-$n$-gram rule & $0.8922$ & $0.79$ & $0.594$ & $0.056$ & $0.303$$^\dagger$ & $+156$$^\dagger$ & $3.2$ & $0.014$ \\
    \quad windowed seq-rep-$n$ & $0.9887$ & $0.98$ & $0.861$ & $0.308$ & $0.364$ & $+205$ & $5.3$ & $0.186$ \\
    \quad rep$/\ell$ & $0.9978$ & $1.00$ & $0.859$ & $0.312$ & $0.359$ & $+252$ & $4.1$ & $0.195$ \\
    \addlinespace[2pt]
    \quad windowed entropy & $0.9674$ & $0.84$ & $0.371$ & $0.025$ & $0.122$$^\dagger$ & $+951$$^\dagger$ & $5.2$ & $0.034$ \\
    \quad \begin{tabular}[t]{@{}l@{}}windowed negative\\ log-probability\end{tabular} & $0.9704$ & $0.95$ & $0.466$ & $0.018$ & $0.102$ & $+924$ & $4.6$ & $0.028$ \\
    \addlinespace[2pt]
    \quad frozen probe, L15 & $0.9988\sd{0.0002}$ & $1.00\sd{0.00}$ & $0.628\sd{0.013}$ & $0.087\sd{0.008}$ & $0.603\sd{0.011}$ & $-144\sd{21}$ & $5.2\sd{0.1}$ & $0.029\sd{0.004}$ \\
    \quad adapted probe, L15 & $0.9983\sd{0.0005}$ & $1.00\sd{0.00}$ & $\mathbf{0.943\sd{0.008}}$ & $\mathbf{0.487\sd{0.009}}$ & $\mathbf{0.653\sd{0.019}}$ & $\mathbf{-238\sd{6}}$ & $5.3\sd{0.2}$ & $0.239\sd{0.006}$ \\
    \quad adapted probe, L6 & $0.9955\sd{0.0013}$ & $1.00\sd{0.00}$ & $0.922\sd{0.018}$ & $0.458\sd{0.017}$ & $0.580\sd{0.021}$ & $-179\sd{41}$ & $5.1\sd{0.2}$ & $0.229\sd{0.002}$ \\
    \midrule
    \multicolumn{9}{@{}l}{\textsc{Mistral} ($n=318$)} \\
    \quad random alarms & $0.9029$ & $0.35$ & $1\times10^{-4}$ & $1\times10^{-4}$ & $0.167$$^\dagger$ & $+1256$$^\dagger$ & $4.8$ & $1\times10^{-4}$ \\
    \quad token-cap rule & $\mathbf{1.0000}$ & $1.00$ & $3\times10^{-4}$ & $0.000$ & $0.000$ & $+3691$ & $0.0$ & $0.000$ \\
    \addlinespace[2pt]
    \quad repeated-$n$-gram rule & $0.9493$ & $0.90$ & $0.749$ & $0.008$ & $0.070$$^\dagger$ & $+158$$^\dagger$ & $1.8$ & $0.007$ \\
    \quad windowed seq-rep-$n$ & $0.9981$ & $1.00$ & $\mathbf{0.933}$ & \underline{$0.190$} & $0.204$ & $+196$ & $5.5$ & $0.193$ \\
    \quad rep$/\ell$ & $0.9979$ & $1.00$ & $0.916$ & \underline{$0.166$} & $0.201$ & $+298$ & $4.4$ & $0.193$ \\
    \addlinespace[2pt]
    \quad windowed entropy & $0.9927$ & $0.97$ & $0.814$ & $0.037$ & $0.077$ & $+496$ & $4.5$ & $0.057$ \\
    \quad \begin{tabular}[t]{@{}l@{}}windowed negative\\ log-probability\end{tabular} & $0.9921$ & $0.98$ & $0.810$ & $0.021$ & $0.054$ & $+351$ & $3.2$ & $0.038$ \\
    \addlinespace[2pt]
    \quad frozen probe, L5 & $0.9919\sd{0.0003}$ & $0.99\sd{0.00}$ & $0.570\sd{0.018}$ & $0.033\sd{0.002}$ & $0.261\sd{0.009}$ & $+112\sd{7}$ & $5.1\sd{0.1}$ & $0.047\sd{0.002}$ \\
    \quad adapted probe, L5 & $0.9944\sd{0.0017}$ & $0.99\sd{0.01}$ & $0.859\sd{0.054}$ & \underline{$0.152\sd{0.017}$} & $0.306\sd{0.011}$ & $+79\sd{3}$ & $5.1\sd{0.5}$ & $0.140\sd{0.013}$ \\
    \quad adapted probe, L11 & $0.9951\sd{0.0006}$ & $0.99\sd{0.00}$ & $0.808\sd{0.044}$ & $\mathbf{0.223\sd{0.081}}$ & $\mathbf{0.392\sd{0.055}}$ & $\mathbf{+46\sd{14}}$ & $4.4\sd{0.6}$ & $0.213\sd{0.072}$ \\
    \bottomrule
  \end{tabular}%
  }
\end{table}

\begin{table}[!htbp]
  \centering
  \caption{As \Cref{tab:baselines}, on the held-out sources (\texttt{codeforces} and \texttt{medical\_o1}), with judge labels.}
  \label{tab:heldout}
  \footnotesize
  \setlength{\tabcolsep}{3pt}
  \resizebox{\linewidth}{!}{%
  \begin{tabular}{@{}lrrrrrrrr@{}}
    \toprule
    & \shortstack{Response\\classification} & \multicolumn{2}{c}{Failure detection} & \multicolumn{3}{c}{Early warning} & & \\
    \cmidrule(lr){2-2}\cmidrule(lr){3-4}\cmidrule(lr){5-7}
    Scorer & AUROC $\uparrow$ & Detection $\uparrow$ & Post $\uparrow$ & $\mathrm{Warn}_{256}$ $\uparrow$ & Early $\uparrow$ & Lead $\downarrow$ & FA (\%) & Pre \\
    \midrule
    \multicolumn{9}{@{}l}{\textsc{Apertus} ($n=26$)} \\
    \quad random alarms & $0.9150$ & $0.32$ & $1\times10^{-4}$ & $1\times10^{-4}$ & $0.244$$^\dagger$ & $+986$$^\dagger$ & $3.7$ & $1\times10^{-4}$ \\
    \quad token-cap rule & $\mathbf{1.0000}$ & $1.00$ & $3\times10^{-4}$ & $0.000$ & $0.000$ & $+3440$ & $0.0$ & $0.000$ \\
    \addlinespace[2pt]
    \quad repeated-$n$-gram rule & $0.9137$ & $0.85$ & $0.646$ & $0.017$ & $0.227$$^\dagger$ & $+149$$^\dagger$ & $0.8$ & $0.002$ \\
    \quad windowed seq-rep-$n$ & $0.9992$ & $0.92$ & $0.843$ & $0.069$ & $0.083$ & $+268$ & $0.6$ & $0.005$ \\
    \quad rep$/\ell$ & $\mathbf{1.0000}$ & $1.00$ & $0.893$ & $0.260$ & $0.346$ & $+373$ & $1.5$ & $0.097$ \\
    \addlinespace[2pt]
    \quad windowed entropy & $0.9998$ & $1.00$ & $0.777$ & $0.016$ & $0.038$ & $+530$ & $1.2$ & $0.006$ \\
    \quad \begin{tabular}[t]{@{}l@{}}windowed negative\\ log-probability\end{tabular} & $0.9995$ & $1.00$ & $0.803$ & $0.042$ & $0.038$ & $+383$ & $1.4$ & $0.025$ \\
    \addlinespace[2pt]
    \quad frozen probe, L15 & $\mathbf{1.0000\sd{0.0000}}$ & $1.00\sd{0.00}$ & $0.790\sd{0.014}$ & $0.287\sd{0.003}$ & $0.577\sd{0.000}$ & $-68\sd{19}$ & $0.3\sd{0.0}$ & $0.091\sd{0.002}$ \\
    \quad adapted probe, L15 & $\mathbf{1.0000\sd{0.0000}}$ & $1.00\sd{0.00}$ & $0.999\sd{0.000}$ & $0.802\sd{0.030}$ & $\mathbf{0.987\sd{0.022}}$ & $\mathbf{-633\sd{22}}$ & $0.8\sd{0.4}$ & $0.859\sd{0.033}$ \\
    \quad adapted probe, L30 & $\mathbf{1.0000\sd{0.0000}}$ & $1.00\sd{0.00}$ & $\mathbf{1.000\sd{0.000}}$ & $\mathbf{0.916\sd{0.048}}$ & $\mathbf{0.987\sd{0.022}}$ & \underline{$-632\sd{24}$} & $0.2\sd{0.1}$ & $0.938\sd{0.034}$ \\
    \midrule
    \multicolumn{9}{@{}l}{\textsc{Llama} ($n=133$)} \\
    \quad random alarms & $0.8998$ & $0.29$ & $9\times10^{-5}$ & $9\times10^{-5}$ & $0.142$$^\dagger$ & $+1371$$^\dagger$ & $4.0$ & $9\times10^{-5}$ \\
    \quad token-cap rule & $\mathbf{1.0000}$ & $1.00$ & $3\times10^{-4}$ & $0.000$ & $0.000$ & $+3762$ & $0.0$ & $0.000$ \\
    \addlinespace[2pt]
    \quad repeated-$n$-gram rule & $0.9615$ & $0.92$ & $0.770$ & $0.037$ & $0.114$ & $+211$ & $1.6$ & $0.015$ \\
    \quad windowed seq-rep-$n$ & $0.9980$ & $0.97$ & $0.887$ & $0.106$ & $0.109$ & $+220$ & $0.5$ & $0.213$ \\
    \quad rep$/\ell$ & $0.9999$ & $0.99$ & $0.817$ & $0.072$ & $0.068$ & $+562$ & $0.3$ & $0.163$ \\
    \addlinespace[2pt]
    \quad windowed entropy & $0.9849$ & $0.80$ & $0.400$ & $0.020$ & $0.037$$^\dagger$ & $+1018$$^\dagger$ & $2.2$ & $0.010$ \\
    \quad \begin{tabular}[t]{@{}l@{}}windowed negative\\ log-probability\end{tabular} & $0.9838$ & $0.90$ & $0.524$ & $0.016$ & $0.025$ & $+874$ & $2.3$ & $0.008$ \\
    \addlinespace[2pt]
    \quad frozen probe, L15 & $\mathbf{1.0000\sd{0.0000}}$ & $1.00\sd{0.00}$ & $0.715\sd{0.015}$ & $0.032\sd{0.004}$ & $0.198\sd{0.009}$ & $+84\sd{8}$ & $0.3\sd{0.0}$ & $0.023\sd{0.003}$ \\
    \quad adapted probe, L15 & $0.9999\sd{0.0000}$ & $1.00\sd{0.00}$ & $\mathbf{0.964\sd{0.004}}$ & $\mathbf{0.604\sd{0.042}}$ & $\mathbf{0.559\sd{0.004}}$ & $\mathbf{-155\sd{12}}$ & $0.3\sd{0.1}$ & $0.755\sd{0.075}$ \\
    \quad adapted probe, L6 & $0.9999\sd{0.0000}$ & $1.00\sd{0.00}$ & \underline{$0.915\sd{0.050}$} & \underline{$0.597\sd{0.012}$} & $0.530\sd{0.004}$ & $-86\sd{55}$ & $0.1\sd{0.0}$ & $0.748\sd{0.007}$ \\
    \midrule
    \multicolumn{9}{@{}l}{\textsc{Mistral} ($n=132$)} \\
    \quad random alarms & $0.9120$ & $0.35$ & $1\times10^{-4}$ & $1\times10^{-4}$ & $0.194$$^\dagger$ & $+1162$$^\dagger$ & $4.3$ & $1\times10^{-4}$ \\
    \quad token-cap rule & $\mathbf{1.0000}$ & $1.00$ & $3\times10^{-4}$ & $0.000$ & $0.000$ & $+3594$ & $0.0$ & $0.000$ \\
    \addlinespace[2pt]
    \quad repeated-$n$-gram rule & $0.9485$ & $0.89$ & $0.703$ & $0.033$ & $0.128$$^\dagger$ & $+169$$^\dagger$ & $0.6$ & $0.027$ \\
    \quad windowed seq-rep-$n$ & $0.9998$ & $1.00$ & \underline{$0.881$} & $0.116$ & $0.144$ & $+334$ & $0.8$ & $0.166$ \\
    \quad rep$/\ell$ & $0.9999$ & $1.00$ & \underline{$0.862$} & $0.109$ & $0.114$ & $+492$ & $0.7$ & $0.168$ \\
    \addlinespace[2pt]
    \quad windowed entropy & $0.9962$ & $1.00$ & \underline{$0.819$} & $0.072$ & $0.129$ & $+502$ & $6.0$ & $0.137$ \\
    \quad \begin{tabular}[t]{@{}l@{}}windowed negative\\ log-probability\end{tabular} & $0.9938$ & $1.00$ & \underline{$0.790$} & $0.043$ & $0.152$ & $+338$ & $4.0$ & $0.127$ \\
    \addlinespace[2pt]
    \quad frozen probe, L5 & $0.9990\sd{0.0002}$ & $0.96\sd{0.01}$ & $0.352\sd{0.033}$ & $0.026\sd{0.003}$ & $0.113\sd{0.018}$ & $+270\sd{24}$ & $0.3\sd{0.1}$ & $0.032\sd{0.009}$ \\
    \quad adapted probe, L5 & $0.9997\sd{0.0001}$ & $0.98\sd{0.00}$ & \underline{$0.800\sd{0.103}$} & $0.125\sd{0.027}$ & $0.305\sd{0.016}$ & $+90\sd{11}$ & $0.3\sd{0.1}$ & $0.121\sd{0.057}$ \\
    \quad adapted probe, L11 & \underline{$0.9998\sd{0.0003}$} & $1.00\sd{0.00}$ & $\mathbf{0.885\sd{0.143}}$ & $\mathbf{0.632\sd{0.487}}$ & $\mathbf{0.720\sd{0.399}}$ & $\mathbf{-296\sd{335}}$ & $0.5\sd{0.2}$ & $0.652\sd{0.490}$ \\
    \bottomrule
  \end{tabular}%
  }
\end{table}

\begin{table}[!htbp]
  \centering
  \caption{As \Cref{tab:heldout}, with human-verified labels.}
  \label{tab:heldout-human}
  \footnotesize
  \setlength{\tabcolsep}{3pt}
  \resizebox{\linewidth}{!}{%
  \begin{tabular}{@{}lrrrrrrrr@{}}
    \toprule
    & \shortstack{Response\\classification} & \multicolumn{2}{c}{Failure detection} & \multicolumn{3}{c}{Early warning} & & \\
    \cmidrule(lr){2-2}\cmidrule(lr){3-4}\cmidrule(lr){5-7}
    Scorer & AUROC $\uparrow$ & Detection $\uparrow$ & Post $\uparrow$ & $\mathrm{Warn}_{256}$ $\uparrow$ & Early $\uparrow$ & Lead $\downarrow$ & FA (\%) & Pre \\
    \midrule
    \multicolumn{9}{@{}l}{\textsc{Apertus} ($n=24$)} \\
    \quad random alarms & $0.9149$ & $0.32$ & $1\times10^{-4}$ & $1\times10^{-4}$ & $0.218$$^\dagger$ & $+1081$$^\dagger$ & $3.7$ & $1\times10^{-4}$ \\
    \quad token-cap rule & $0.9999$ & $1.00$ & $3\times10^{-4}$ & $0.000$ & $0.000$ & $+3440$ & $0.0$ & $0.000$ \\
    \addlinespace[2pt]
    \quad repeated-$n$-gram rule & $0.9532$ & $0.92$ & $0.680$ & $0.001$ & $0.136$ & $+190$ & $0.8$ & $0.001$ \\
    \quad windowed seq-rep-$n$ & $0.9991$ & $0.92$ & $0.805$ & $0.028$ & $0.045$ & $+474$ & $0.6$ & $0.000$ \\
    \quad rep$/\ell$ & $\mathbf{1.0000}$ & $1.00$ & $0.866$ & $0.196$ & $0.250$ & $+430$ & $1.5$ & $0.068$ \\
    \addlinespace[2pt]
    \quad windowed entropy & $0.9999$ & $1.00$ & $0.758$ & $0.017$ & $0.042$ & $+578$ & $1.2$ & $0.008$ \\
    \quad \begin{tabular}[t]{@{}l@{}}windowed negative\\ log-probability\end{tabular} & $0.9997$ & $1.00$ & $0.798$ & $0.044$ & $0.042$ & $+427$ & $1.4$ & $0.032$ \\
    \addlinespace[2pt]
    \quad frozen probe, L15 & $\mathbf{1.0000\sd{0.0000}}$ & $1.00\sd{0.00}$ & $0.822\sd{0.010}$ & $0.164\sd{0.002}$ & $0.458\sd{0.000}$ & $+33\sd{8}$ & $0.3\sd{0.0}$ & $0.047\sd{0.002}$ \\
    \quad adapted probe, L15 & $0.9999\sd{0.0000}$ & $1.00\sd{0.00}$ & $0.999\sd{0.000}$ & $0.796\sd{0.026}$ & \underline{$0.986\sd{0.024}$} & $\mathbf{-633\sd{22}}$ & $0.8\sd{0.4}$ & $0.823\sd{0.042}$ \\
    \quad adapted probe, L30 & $0.9999\sd{0.0000}$ & $1.00\sd{0.00}$ & $\mathbf{1.000\sd{0.000}}$ & $\mathbf{0.917\sd{0.048}}$ & $\mathbf{1.000\sd{0.000}}$ & \underline{$-632\sd{24}$} & $0.2\sd{0.1}$ & $0.923\sd{0.042}$ \\
    \midrule
    \multicolumn{9}{@{}l}{\textsc{Llama} ($n=133$)} \\
    \quad random alarms & $0.8998$ & $0.29$ & $9\times10^{-5}$ & $9\times10^{-5}$ & $0.142$$^\dagger$ & $+1370$$^\dagger$ & $4.0$ & $9\times10^{-5}$ \\
    \quad token-cap rule & $\mathbf{1.0000}$ & $1.00$ & $3\times10^{-4}$ & $0.000$ & $0.000$ & $+3762$ & $0.0$ & $0.000$ \\
    \addlinespace[2pt]
    \quad repeated-$n$-gram rule & $0.9615$ & $0.92$ & $0.770$ & $0.037$ & $0.122$ & $+210$ & $1.6$ & $0.015$ \\
    \quad windowed seq-rep-$n$ & $0.9980$ & $0.97$ & $0.888$ & $0.105$ & $0.109$ & $+219$ & $0.5$ & $0.210$ \\
    \quad rep$/\ell$ & $0.9999$ & $0.99$ & $0.817$ & $0.068$ & $0.061$ & $+560$ & $0.3$ & $0.163$ \\
    \addlinespace[2pt]
    \quad windowed entropy & $0.9849$ & $0.80$ & $0.401$ & $0.020$ & $0.037$$^\dagger$ & $+1018$$^\dagger$ & $2.2$ & $0.011$ \\
    \quad \begin{tabular}[t]{@{}l@{}}windowed negative\\ log-probability\end{tabular} & $0.9838$ & $0.90$ & $0.524$ & $0.016$ & $0.025$ & $+874$ & $2.3$ & $0.008$ \\
    \addlinespace[2pt]
    \quad frozen probe, L15 & $\mathbf{1.0000\sd{0.0000}}$ & $1.00\sd{0.00}$ & $0.716\sd{0.015}$ & $0.032\sd{0.004}$ & $0.201\sd{0.011}$ & $+83\sd{9}$ & $0.3\sd{0.0}$ & $0.021\sd{0.003}$ \\
    \quad adapted probe, L15 & $0.9999\sd{0.0000}$ & $1.00\sd{0.00}$ & $\mathbf{0.965\sd{0.004}}$ & $\mathbf{0.597\sd{0.041}}$ & $\mathbf{0.559\sd{0.004}}$ & $\mathbf{-155\sd{12}}$ & $0.3\sd{0.1}$ & $0.754\sd{0.076}$ \\
    \quad adapted probe, L6 & $0.9999\sd{0.0000}$ & $1.00\sd{0.00}$ & \underline{$0.916\sd{0.050}$} & \underline{$0.590\sd{0.012}$} & $0.530\sd{0.004}$ & $-86\sd{55}$ & $0.1\sd{0.0}$ & $0.747\sd{0.007}$ \\
    \midrule
    \multicolumn{9}{@{}l}{\textsc{Mistral} ($n=129$)} \\
    \quad random alarms & $0.9119$ & $0.35$ & $1\times10^{-4}$ & $1\times10^{-4}$ & $0.196$$^\dagger$ & $+1153$$^\dagger$ & $4.3$ & $1\times10^{-4}$ \\
    \quad token-cap rule & $\mathbf{0.9999}$ & $1.00$ & $3\times10^{-4}$ & $0.000$ & $0.000$ & $+3568$ & $0.0$ & $0.000$ \\
    \addlinespace[2pt]
    \quad repeated-$n$-gram rule & $0.9594$ & $0.91$ & $0.721$ & $0.034$ & $0.128$ & $+167$ & $0.6$ & $0.027$ \\
    \quad windowed seq-rep-$n$ & $0.9997$ & $1.00$ & \underline{$0.885$} & $0.119$ & $0.147$ & $+322$ & $0.8$ & $0.165$ \\
    \quad rep$/\ell$ & $0.9998$ & $1.00$ & \underline{$0.863$} & $0.112$ & $0.116$ & $+488$ & $0.7$ & $0.167$ \\
    \addlinespace[2pt]
    \quad windowed entropy & $0.9961$ & $1.00$ & \underline{$0.824$} & $0.077$ & $0.132$ & $+504$ & $6.0$ & $0.136$ \\
    \quad \begin{tabular}[t]{@{}l@{}}windowed negative\\ log-probability\end{tabular} & $0.9936$ & $1.00$ & \underline{$0.797$} & $0.047$ & $0.155$ & $+334$ & $4.1$ & $0.126$ \\
    \addlinespace[2pt]
    \quad frozen probe, L5 & $0.9989\sd{0.0002}$ & $0.96\sd{0.01}$ & $0.352\sd{0.033}$ & $0.027\sd{0.003}$ & $0.115\sd{0.018}$ & $+258\sd{27}$ & $0.4\sd{0.1}$ & $0.031\sd{0.009}$ \\
    \quad adapted probe, L5 & $0.9996\sd{0.0002}$ & $0.98\sd{0.00}$ & \underline{$0.806\sd{0.103}$} & $0.130\sd{0.029}$ & $0.318\sd{0.018}$ & $+79\sd{15}$ & $0.3\sd{0.1}$ & $0.120\sd{0.057}$ \\
    \quad adapted probe, L11 & \underline{$0.9997\sd{0.0004}$} & $1.00\sd{0.00}$ & $\mathbf{0.888\sd{0.143}}$ & $\mathbf{0.643\sd{0.495}}$ & $\mathbf{0.727\sd{0.406}}$ & $\mathbf{-316\sd{345}}$ & $0.5\sd{0.2}$ & $0.654\sd{0.491}$ \\
    \bottomrule
  \end{tabular}%
  }
\end{table}

\begin{table}[!htbp]
\centering
\caption{Warning coverage and early-warning rate per source, for the frozen probe and the adapted probe at its own
selected layer, at $\beta=5\%$ with the thresholds calibrated on the whole validation split. The
first five sources form the in-domain test split, the last two the held-out split. $n$: number of degenerate
responses of the source. Rates are left out (--) for sources with fewer than $10$ degenerate
responses. Human-verified labels, mean over three seeds.}
\label{tab:per-source}
\setlength{\tabcolsep}{3pt}
\resizebox{\linewidth}{!}{%
\begin{tabular}{l rcc rcc rcc}
\toprule
& \multicolumn{3}{c}{\textsc{Apertus}} & \multicolumn{3}{c}{\textsc{Llama}} & \multicolumn{3}{c}{\textsc{Mistral}} \\
\cmidrule(lr){2-4} \cmidrule(lr){5-7} \cmidrule(lr){8-10}
Source & $n$ & frozen, L15 & adapted, L30 & $n$ & frozen, L15 & adapted, L6 & $n$ & frozen, L5 & adapted, L11 \\
\midrule
\multicolumn{10}{l}{\textbf{$\mathrm{Warn}_{256}$}} \\
\texttt{deepmath\_103k} & 23 & 0.07 & 0.38 & 103 & 0.08 & 0.42 & 123 & 0.04 & 0.18 \\
\texttt{numinamath\_1\_5} & 33 & 0.23 & 0.63 & 109 & 0.11 & 0.47 & 100 & 0.03 & 0.17 \\
\texttt{aime\_2025} & 1 & -- & -- & 16 & 0.09 & 0.62 & 22 & 0.03 & 0.43 \\
\texttt{if\_sft\_data\_verified} & 37 & 0.18 & 0.46 & 33 & 0.04 & 0.28 & 40 & 0.03 & 0.13 \\
\texttt{llama\_nemotron} & 22 & 0.25 & 0.89 & 23 & 0.04 & 0.71 & 30 & 0.00 & 0.48 \\
\addlinespace[2pt]
\texttt{codeforces} & 23 & 0.17 & 0.95 & 73 & 0.04 & 0.84 & 128 & 0.02 & 0.64 \\
\texttt{medical\_o1} & 1 & -- & -- & 60 & 0.02 & 0.00 & 1 & -- & -- \\
\midrule
\multicolumn{10}{l}{\textbf{Early}} \\
\texttt{deepmath\_103k} & 23 & 0.62 & 0.64 & 103 & 0.67 & 0.55 & 123 & 0.30 & 0.37 \\
\texttt{numinamath\_1\_5} & 33 & 0.81 & 0.82 & 109 & 0.65 & 0.60 & 100 & 0.26 & 0.35 \\
\texttt{aime\_2025} & 1 & -- & -- & 16 & 0.77 & 0.77 & 22 & 0.53 & 0.70 \\
\texttt{if\_sft\_data\_verified} & 37 & 0.51 & 0.53 & 33 & 0.26 & 0.34 & 40 & 0.19 & 0.30 \\
\texttt{llama\_nemotron} & 22 & 0.70 & 0.98 & 23 & 0.33 & 0.79 & 30 & 0.01 & 0.56 \\
\addlinespace[2pt]
\texttt{codeforces} & 23 & 0.43 & 1.00 & 73 & 0.26 & 0.95 & 128 & 0.11 & 0.72 \\
\texttt{medical\_o1} & 1 & -- & -- & 60 & 0.13 & 0.02 & 1 & -- & -- \\
\bottomrule
\end{tabular}%
}
\end{table}

\FloatBarrier
\subsection{\texorpdfstring{Performance at $\beta=1\%$ and $\beta=10\%$}{Performance at beta = 1\% and beta = 10\%}}
\label{app:budgets}

\Cref{tab:budget-1pct,tab:budget-10pct} report the same comparison as
\Cref{tab:baselines} at $\beta=1\%$ and $\beta=10\%$, on the human-verified
in-domain test split.

\begin{table}[!htbp]
  \centering
  \caption{As \Cref{tab:baselines}, at $\beta=1\%$.}
  \label{tab:budget-1pct}
  \footnotesize
  \setlength{\tabcolsep}{3pt}
  \resizebox{\linewidth}{!}{%
  \begin{tabular}{@{}lrrrrrrrr@{}}
    \toprule
    & \shortstack{Response\\classification} & \multicolumn{2}{c}{Failure detection} & \multicolumn{3}{c}{Early warning} & & \\
    \cmidrule(lr){2-2}\cmidrule(lr){3-4}\cmidrule(lr){5-7}
    Scorer & AUROC $\uparrow$ & Detection $\uparrow$ & Post $\uparrow$ & $\mathrm{Warn}_{256}$ $\uparrow$ & Early $\uparrow$ & Lead $\downarrow$ & FA (\%) & Pre \\
    \midrule
    \multicolumn{9}{@{}l}{\textsc{Apertus} ($n=116$)} \\
    \quad random alarms & $0.8901$ & $0.07$ & $2\times10^{-5}$ & $2\times10^{-5}$ & $0.222$$^\dagger$ & $+1133$$^\dagger$ & $1.0$ & $2\times10^{-5}$ \\
    \quad token-cap rule & $0.9990$ & $1.00$ & $3\times10^{-4}$ & $0.000$ & $0.000$ & $+3276$ & $0.2$ & $0.000$ \\
    \addlinespace[2pt]
    \quad repeated-$n$-gram rule & $0.9972$ & $0.98$ & $0.716$ & $0.002$ & $0.088$ & $+327$ & $1.5$ & $0.001$ \\
    \quad windowed seq-rep-$n$ & $0.9815$ & $0.90$ & \underline{$0.802$} & $0.078$ & $0.115$$^\dagger$ & $+404$$^\dagger$ & $1.2$ & $0.051$ \\
    \quad rep$/\ell$ & $0.9983$ & $0.98$ & \underline{$0.799$} & $0.090$ & $0.105$ & $+592$ & $1.0$ & $0.074$ \\
    \addlinespace[2pt]
    \quad windowed entropy & $0.9971$ & $0.97$ & $0.639$ & $0.000$ & $0.018$ & $+927$ & $0.9$ & $0.023$ \\
    \quad \begin{tabular}[t]{@{}l@{}}windowed negative\\ log-probability\end{tabular} & $0.9964$ & $0.97$ & $0.637$ & $0.004$ & $0.035$ & $+792$ & $1.2$ & $0.024$ \\
    \addlinespace[2pt]
    \quad frozen probe, L15 & $\mathbf{0.9994\sd{0.0000}}$ & $0.98\sd{0.00}$ & $0.709\sd{0.008}$ & $0.056\sd{0.004}$ & $\mathbf{0.428\sd{0.025}}$ & $\mathbf{+38\sd{8}}$ & $1.4\sd{0.1}$ & $0.009\sd{0.001}$ \\
    \quad adapted probe, L15 & $0.9991\sd{0.0001}$ & $0.99\sd{0.00}$ & $\mathbf{0.846\sd{0.057}}$ & $0.220\sd{0.003}$ & \underline{$0.426\sd{0.026}$} & \underline{$+41\sd{19}$} & $1.1\sd{0.1}$ & $0.111\sd{0.010}$ \\
    \quad adapted probe, L30 & \underline{$0.9982\sd{0.0012}$} & $0.95\sd{0.05}$ & \underline{$0.806\sd{0.128}$} & $\mathbf{0.277\sd{0.016}}$ & $\mathbf{0.428\sd{0.022}}$ & \underline{$+44\sd{26}$} & $0.9\sd{0.2}$ & $0.153\sd{0.019}$ \\
    \midrule
    \multicolumn{9}{@{}l}{\textsc{Llama} ($n=284$)} \\
    \quad random alarms & $0.8810$ & $0.07$ & $2\times10^{-5}$ & $2\times10^{-5}$ & $0.315$$^\dagger$ & $+754$$^\dagger$ & $1.0$ & $2\times10^{-5}$ \\
    \quad token-cap rule & $\mathbf{0.9988}$ & $1.00$ & $4\times10^{-4}$ & $0.000$ & $0.000$ & $+2958$ & $0.2$ & $0.000$ \\
    \addlinespace[2pt]
    \quad repeated-$n$-gram rule & $0.8987$ & $0.76$ & $0.527$ & $0.010$ & $0.084$$^\dagger$ & $+414$$^\dagger$ & $0.7$ & $0.003$ \\
    \quad windowed seq-rep-$n$ & $0.9953$ & $0.90$ & $\mathbf{0.644}$ & $\mathbf{0.096}$ & $0.113$ & $+709$ & $1.1$ & $0.067$ \\
    \quad rep$/\ell$ & $0.9981$ & $0.96$ & $0.602$ & $0.081$ & $0.095$ & $+1176$ & $1.2$ & $0.033$ \\
    \addlinespace[2pt]
    \quad windowed entropy & $0.9655$ & $0.19$ & $0.031$ & $0.004$ & $0.145$$^\dagger$ & $+1316$$^\dagger$ & $1.7$ & $0.011$ \\
    \quad \begin{tabular}[t]{@{}l@{}}windowed negative\\ log-probability\end{tabular} & $0.9685$ & $0.13$ & $0.019$ & $0.004$ & $0.211$$^\dagger$ & $+1235$$^\dagger$ & $1.8$ & $0.011$ \\
    \addlinespace[2pt]
    \quad frozen probe, L15 & $0.9986\sd{0.0001}$ & $0.97\sd{0.01}$ & $0.281\sd{0.006}$ & $0.006\sd{0.000}$ & $0.184\sd{0.014}$ & $\mathbf{+250\sd{12}}$ & $1.2\sd{0.2}$ & $0.002\sd{0.000}$ \\
    \quad adapted probe, L15 & \underline{$0.9980\sd{0.0008}$} & $0.97\sd{0.02}$ & $0.539\sd{0.104}$ & $0.049\sd{0.014}$ & $\mathbf{0.243\sd{0.005}}$ & $+306\sd{14}$ & $1.1\sd{0.1}$ & $0.030\sd{0.006}$ \\
    \quad adapted probe, L6 & $0.9948\sd{0.0013}$ & $0.85\sd{0.08}$ & $0.462\sd{0.102}$ & \underline{$0.072\sd{0.029}$} & $0.224\sd{0.026}$$^\dagger$ & $+414\sd{30}$$^\dagger$ & $1.1\sd{0.2}$ & $0.044\sd{0.015}$ \\
    \midrule
    \multicolumn{9}{@{}l}{\textsc{Mistral} ($n=315$)} \\
    \quad random alarms & $0.9026$ & $0.08$ & $2\times10^{-5}$ & $2\times10^{-5}$ & $0.148$$^\dagger$ & $+1420$$^\dagger$ & $1.0$ & $2\times10^{-5}$ \\
    \quad token-cap rule & $\mathbf{0.9995}$ & $1.00$ & $3\times10^{-4}$ & $0.000$ & $0.000$ & $+3688$ & $0.1$ & $0.000$ \\
    \addlinespace[2pt]
    \quad repeated-$n$-gram rule & $0.9539$ & $0.88$ & $0.686$ & $0.003$ & $0.029$$^\dagger$ & $+305$$^\dagger$ & $0.6$ & $0.002$ \\
    \quad windowed seq-rep-$n$ & $0.9981$ & $0.98$ & $\mathbf{0.792}$ & \underline{$0.044$} & $\mathbf{0.055}$ & $\mathbf{+600}$ & $1.1$ & $0.043$ \\
    \quad rep$/\ell$ & $0.9974$ & $1.00$ & $0.751$ & \underline{$0.045$} & $0.044$ & $+921$ & $1.1$ & $0.059$ \\
    \addlinespace[2pt]
    \quad windowed entropy & $0.9928$ & $0.95$ & $0.693$ & $0.005$ & $0.017$ & $+665$ & $1.1$ & $0.017$ \\
    \quad \begin{tabular}[t]{@{}l@{}}windowed negative\\ log-probability\end{tabular} & $0.9922$ & $0.83$ & $0.464$ & $0.003$ & $0.023$$^\dagger$ & $+811$$^\dagger$ & $1.2$ & $0.005$ \\
    \addlinespace[2pt]
    \quad frozen probe, L5 & $0.9916\sd{0.0004}$ & $0.89\sd{0.01}$ & $0.307\sd{0.014}$ & $0.005\sd{0.001}$ & $0.079\sd{0.005}$$^\dagger$ & $+300\sd{10}$$^\dagger$ & $1.5\sd{0.1}$ & $0.007\sd{0.001}$ \\
    \quad adapted probe, L5 & $0.9942\sd{0.0018}$ & $0.88\sd{0.06}$ & $0.534\sd{0.126}$ & \underline{$0.030\sd{0.006}$} & $0.126\sd{0.002}$$^\dagger$ & $+250\sd{31}$$^\dagger$ & $1.1\sd{0.2}$ & $0.040\sd{0.007}$ \\
    \quad adapted probe, L11 & $0.9952\sd{0.0006}$ & $0.88\sd{0.04}$ & $0.326\sd{0.065}$ & $\mathbf{0.075\sd{0.047}}$ & $0.197\sd{0.069}$$^\dagger$ & $+140\sd{38}$$^\dagger$ & $1.1\sd{0.3}$ & $0.069\sd{0.042}$ \\
    \bottomrule
  \end{tabular}%
  }
\end{table}

\begin{table}[!htbp]
  \centering
  \caption{As \Cref{tab:baselines}, at $\beta=10\%$.}
  \label{tab:budget-10pct}
  \footnotesize
  \setlength{\tabcolsep}{3pt}
  \resizebox{\linewidth}{!}{%
  \begin{tabular}{@{}lrrrrrrrr@{}}
    \toprule
    & \shortstack{Response\\classification} & \multicolumn{2}{c}{Failure detection} & \multicolumn{3}{c}{Early warning} & & \\
    \cmidrule(lr){2-2}\cmidrule(lr){3-4}\cmidrule(lr){5-7}
    Scorer & AUROC $\uparrow$ & Detection $\uparrow$ & Post $\uparrow$ & $\mathrm{Warn}_{256}$ $\uparrow$ & Early $\uparrow$ & Lead $\downarrow$ & FA (\%) & Pre \\
    \midrule
    \multicolumn{9}{@{}l}{\textsc{Apertus} ($n=116$)} \\
    \quad random alarms & $0.8901$ & $0.56$ & $2\times10^{-4}$ & $2\times10^{-4}$ & $0.279$$^\dagger$ & $+795$$^\dagger$ & $9.9$ & $2\times10^{-4}$ \\
    \quad token-cap rule & $0.9990$ & $1.00$ & $3\times10^{-4}$ & $0.000$ & $0.000$ & $+3276$ & $0.2$ & $0.000$ \\
    \addlinespace[2pt]
    \quad repeated-$n$-gram rule & $0.9972$ & $0.99$ & $0.850$ & $0.039$ & $0.296$ & $+68$ & $8.6$ & $0.007$ \\
    \quad windowed seq-rep-$n$ & $0.9815$ & $0.94$ & $0.899$ & $0.291$ & $0.413$ & $+40$ & $10.3$ & $0.182$ \\
    \quad rep$/\ell$ & $0.9983$ & $1.00$ & $0.946$ & $0.320$ & $0.405$ & $+76$ & $9.3$ & $0.221$ \\
    \addlinespace[2pt]
    \quad windowed entropy & $0.9971$ & $1.00$ & $0.803$ & $0.038$ & $0.155$ & $+516$ & $10.9$ & $0.066$ \\
    \quad \begin{tabular}[t]{@{}l@{}}windowed negative\\ log-probability\end{tabular} & $0.9964$ & $1.00$ & $0.837$ & $0.030$ & $0.172$ & $+311$ & $11.4$ & $0.062$ \\
    \addlinespace[2pt]
    \quad frozen probe, L15 & $\mathbf{0.9994\sd{0.0000}}$ & $1.00\sd{0.00}$ & $0.905\sd{0.005}$ & $0.281\sd{0.010}$ & $0.750\sd{0.009}$ & $-253\sd{6}$ & $10.7\sd{0.6}$ & $0.111\sd{0.007}$ \\
    \quad adapted probe, L15 & $0.9991\sd{0.0001}$ & $1.00\sd{0.00}$ & $0.985\sd{0.003}$ & $0.658\sd{0.019}$ & $\mathbf{0.799\sd{0.013}}$ & $-340\sd{12}$ & $9.3\sd{0.4}$ & $0.464\sd{0.020}$ \\
    \quad adapted probe, L30 & \underline{$0.9982\sd{0.0012}$} & $1.00\sd{0.00}$ & $\mathbf{0.991\sd{0.002}}$ & $\mathbf{0.700\sd{0.006}}$ & \underline{$0.782\sd{0.020}$} & $\mathbf{-369\sd{26}}$ & $9.8\sd{0.8}$ & $0.499\sd{0.010}$ \\
    \midrule
    \multicolumn{9}{@{}l}{\textsc{Llama} ($n=284$)} \\
    \quad random alarms & $0.8810$ & $0.52$ & $2\times10^{-4}$ & $2\times10^{-4}$ & $0.374$$^\dagger$ & $+472$$^\dagger$ & $9.7$ & $2\times10^{-4}$ \\
    \quad token-cap rule & $\mathbf{0.9988}$ & $1.00$ & $4\times10^{-4}$ & $0.000$ & $0.000$ & $+2958$ & $0.2$ & $0.000$ \\
    \addlinespace[2pt]
    \quad repeated-$n$-gram rule & $0.8987$ & $0.81$ & $0.608$ & $0.056$ & $0.297$$^\dagger$ & $+156$$^\dagger$ & $3.3$ & $0.015$ \\
    \quad windowed seq-rep-$n$ & $0.9953$ & $1.00$ & $0.936$ & $0.464$ & $0.519$ & $-34$ & $9.6$ & $0.295$ \\
    \quad rep$/\ell$ & $0.9981$ & $1.00$ & $0.934$ & $0.482$ & $0.528$ & $-63$ & $8.4$ & $0.327$ \\
    \addlinespace[2pt]
    \quad windowed entropy & $0.9655$ & $0.99$ & $0.663$ & $0.063$ & $0.185$ & $+632$ & $7.8$ & $0.057$ \\
    \quad \begin{tabular}[t]{@{}l@{}}windowed negative\\ log-probability\end{tabular} & $0.9685$ & $0.98$ & $0.644$ & $0.042$ & $0.137$ & $+633$ & $7.1$ & $0.042$ \\
    \addlinespace[2pt]
    \quad frozen probe, L15 & $0.9986\sd{0.0001}$ & $1.00\sd{0.00}$ & $0.810\sd{0.007}$ & $0.222\sd{0.010}$ & $0.721\sd{0.007}$ & $-368\sd{23}$ & $9.6\sd{0.1}$ & $0.092\sd{0.005}$ \\
    \quad adapted probe, L15 & \underline{$0.9980\sd{0.0008}$} & $1.00\sd{0.00}$ & $\mathbf{0.986\sd{0.001}}$ & $\mathbf{0.688\sd{0.001}}$ & $\mathbf{0.790\sd{0.005}}$ & $\mathbf{-503\sd{19}}$ & $10.0\sd{0.2}$ & $0.409\sd{0.008}$ \\
    \quad adapted probe, L6 & $0.9948\sd{0.0013}$ & $1.00\sd{0.00}$ & $0.973\sd{0.006}$ & $0.622\sd{0.008}$ & $0.727\sd{0.011}$ & $-411\sd{6}$ & $9.8\sd{0.4}$ & $0.379\sd{0.006}$ \\
    \midrule
    \multicolumn{9}{@{}l}{\textsc{Mistral} ($n=315$)} \\
    \quad random alarms & $0.9026$ & $0.60$ & $2\times10^{-4}$ & $2\times10^{-4}$ & $0.195$$^\dagger$ & $+1041$$^\dagger$ & $9.6$ & $2\times10^{-4}$ \\
    \quad token-cap rule & $\mathbf{0.9995}$ & $1.00$ & $3\times10^{-4}$ & $0.000$ & $0.000$ & $+3688$ & $0.1$ & $0.000$ \\
    \addlinespace[2pt]
    \quad repeated-$n$-gram rule & $0.9539$ & $0.92$ & $0.801$ & $0.036$ & $0.197$ & $+90$ & $6.6$ & $0.030$ \\
    \quad windowed seq-rep-$n$ & $0.9981$ & $1.00$ & $\mathbf{0.961}$ & \underline{$0.280$} & $0.321$ & $+102$ & $10.1$ & $0.264$ \\
    \quad rep$/\ell$ & $0.9974$ & $1.00$ & $0.948$ & \underline{$0.263$} & $0.295$ & $+168$ & $8.7$ & $0.257$ \\
    \addlinespace[2pt]
    \quad windowed entropy & $0.9928$ & $0.99$ & $0.853$ & $0.065$ & $0.109$ & $+425$ & $8.2$ & $0.082$ \\
    \quad \begin{tabular}[t]{@{}l@{}}windowed negative\\ log-probability\end{tabular} & $0.9922$ & $1.00$ & $0.882$ & $0.045$ & $0.102$ & $+275$ & $8.3$ & $0.064$ \\
    \addlinespace[2pt]
    \quad frozen probe, L5 & $0.9916\sd{0.0004}$ & $0.99\sd{0.00}$ & $0.701\sd{0.015}$ & $0.067\sd{0.003}$ & $0.381\sd{0.006}$ & $+57\sd{3}$ & $9.3\sd{0.3}$ & $0.088\sd{0.002}$ \\
    \quad adapted probe, L5 & $0.9942\sd{0.0018}$ & $1.00\sd{0.00}$ & $0.937\sd{0.016}$ & \underline{$0.252\sd{0.017}$} & $0.449\sd{0.008}$ & $+20\sd{5}$ & $8.7\sd{0.2}$ & $0.220\sd{0.006}$ \\
    \quad adapted probe, L11 & $0.9952\sd{0.0006}$ & $1.00\sd{0.00}$ & $0.908\sd{0.033}$ & $\mathbf{0.309\sd{0.093}}$ & $\mathbf{0.507\sd{0.054}}$ & $\mathbf{-40\sd{52}}$ & $8.7\sd{0.9}$ & $0.307\sd{0.085}$ \\
    \bottomrule
  \end{tabular}%
  }
\end{table}

\FloatBarrier
\subsection{Persistence}
\label{app:persistence}

Requiring $m$ consecutive flagged tokens before an alarm (\Cref{eq:alarm} in
\Cref{app:metrics}) suppresses isolated flags on healthy responses, so at the
same false-alarm budget the calibrated threshold drops as $m$ grows.
\Cref{tab:persistence} reports the probes at $m\in\{1,16,64\}$. Coverage counts
every token at or above the threshold of that $m$, while the first alarm is the
token that completes the run, $m-1$ tokens after its first flag. A larger $m$
raises warning coverage because its lower threshold flags more tokens. Many of
these flags come before a run of $m$ is complete, where the alarm cannot fire
yet, so the first alarm does not come earlier (Lead in \Cref{tab:persistence}).
The main text uses $m=1$.

\begin{table}[!htbp]
  \centering
  \caption{Requiring consecutive flags raises coverage but does not move the
first alarm earlier. Probes at persistence $m \in \{1, 16, 64\}$, the
number of consecutive flagged tokens needed for an alarm (\Cref{eq:alarm}),
with the threshold recalibrated at each $m$ for $\beta = 5\%$. Coverage
counts every token at or above the threshold, while the first alarm is the
token that completes the run. Lead in tokens, negative for warnings.
Human-verified in-domain test split, mean over three seeds.}
  \label{tab:persistence}
  \footnotesize
  \setlength{\tabcolsep}{3.5pt}
  \begin{tabular}{ll ccc ccc ccc ccc}
    \toprule
    & & \multicolumn{3}{c}{Post} & \multicolumn{3}{c}{$\mathrm{Warn}_{256}$} & \multicolumn{3}{c}{Early} & \multicolumn{3}{c}{Lead} \\
    \cmidrule(lr){3-5}\cmidrule(lr){6-8}\cmidrule(lr){9-11}\cmidrule(lr){12-14}
    Model & Probe & $1$ & $16$ & $64$ & $1$ & $16$ & $64$ & $1$ & $16$ & $64$ & $1$ & $16$ & $64$ \\
    \midrule
    \textsc{Apertus} & frozen, L15 & 0.865 & 0.954 & 0.984 & 0.186 & 0.511 & 0.724 & 0.658 & 0.649 & 0.621 & $-172$ & $-89$ & $-105$ \\
     & adapted, L15 & 0.966 & 0.981 & 0.991 & 0.523 & 0.623 & 0.711 & 0.695 & 0.687 & 0.672 & $-214$ & $-185$ & $-177$ \\
     & adapted, L30 & 0.973 & 0.987 & 0.992 & 0.577 & 0.656 & 0.712 & 0.716 & 0.713 & 0.681 & $-232$ & $-216$ & $-204$ \\
    \midrule
    \textsc{Llama} & frozen, L15 & 0.635 & 0.936 & 0.982 & 0.087 & 0.466 & 0.693 & 0.594 & 0.568 & 0.561 & $-137$ & $-99$ & $-108$ \\
     & adapted, L15 & 0.945 & 0.974 & 0.985 & 0.488 & 0.608 & 0.676 & 0.655 & 0.653 & 0.641 & $-236$ & $-231$ & $-224$ \\
     & adapted, L6 & 0.920 & 0.954 & 0.967 & 0.462 & 0.551 & 0.597 & 0.578 & 0.568 & 0.572 & $-182$ & $-171$ & $-152$ \\
    \midrule
    \textsc{Mistral} & frozen, L5 & 0.572 & 0.887 & 0.961 & 0.031 & 0.214 & 0.408 & 0.262 & 0.250 & 0.259 & $+108$ & $+110$ & $+114$ \\
     & adapted, L5 & 0.862 & 0.940 & 0.967 & 0.147 & 0.255 & 0.359 & 0.309 & 0.311 & 0.296 & $+76$ & $+78$ & $+103$ \\
     & adapted, L11 & 0.813 & 0.927 & 0.965 & 0.220 & 0.342 & 0.446 & 0.393 & 0.383 & 0.371 & $+44$ & $+48$ & $+80$ \\
    \bottomrule
  \end{tabular}
\end{table}

\FloatBarrier
\subsection{Representations before and after adaptation}
\label{app:pca}

\begin{figure}[h]
  \centering
  \includegraphics[width=\linewidth]{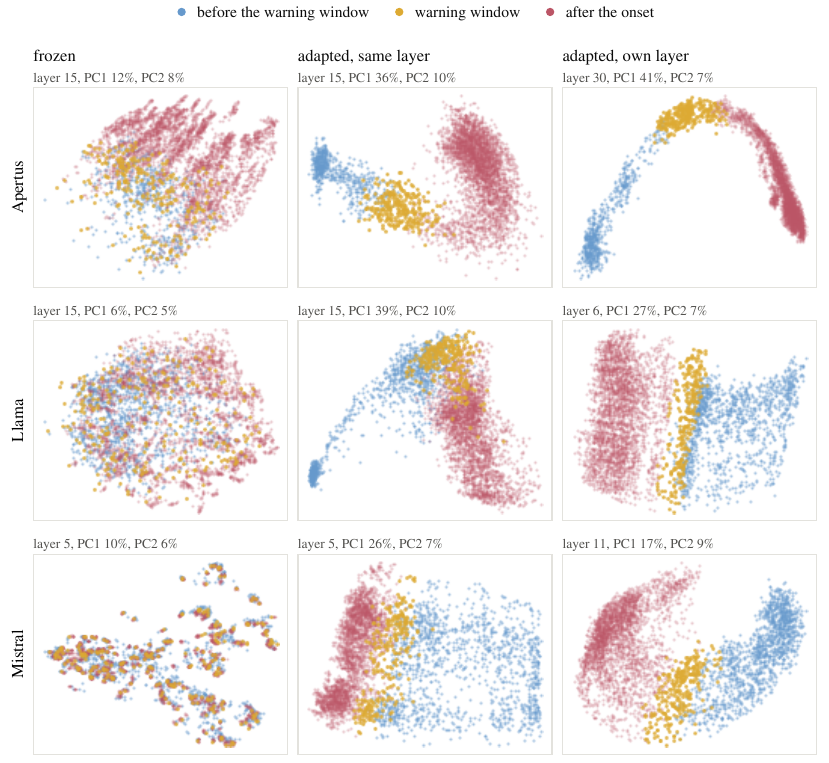}
  \caption{In the frozen model, the tokens before the warning window, in it
and after the onset overlap; after adaptation they fall in largely separate
regions. Residual-stream activations of the tokens of one degenerate
response per model (\textsc{Apertus}: \texttt{numinamath\_1\_5\_00388},
\textsc{Llama}: \texttt{numinamath\_1\_5\_00044}, \textsc{Mistral}:
\texttt{numinamath\_1\_5\_00517}, all from the in-domain test split, projected onto their first two
principal components (x: PC1, y: PC2). Columns: the layer read by the
frozen probe (left), the adapted probe at the same layer (middle) and the
adapted probe at its own selected layer (right). Each panel has its own
projection, and its title gives the share of variance each component
explains. Blue: before the warning window. Yellow: warning window. Red:
after the onset. The adapted panels use the seed with the highest
validation warning coverage.}
  \label{fig:pca-rollouts}
\end{figure}

\Cref{fig:pca-rollouts} follows one degenerate response per model through the
three probes of \Cref{sec:rq-adaptation}. For each probe we take the residual
stream of the response at the layer the probe reads, with that probe's adapters
for the two adapted ones, and project every token onto the first two principal
components of the response. In the frozen model, the tokens before the warning
window, in it and after the onset overlap in this plane. After adaptation they
fall in largely separate regions, most clearly at the adapted probe's own layer,
with the warning window between the other two. The two components also explain
more of the variance after adaptation, $26$--$49\%$ against $11$--$20\%$, so the
split between the three regions becomes one of the main directions of variation.
This is the intuition behind the gain in warning coverage of
\Cref{sec:rq-adaptation}.

\end{document}